\documentclass[10pt,twocolumn,letterpaper]{article}

\usepackage[pagenumbers]{cvpr} 

\usepackage{comment}

\definecolor{cvprblue}{rgb}{0.21,0.49,0.74}
\usepackage[pagebackref,breaklinks,colorlinks,allcolors=cvprblue]{hyperref}

\def\paperID{*****} 
\def\confName{CVPR}
\def\confYear{2026}

\newcommand{\metrictablebest}[1]{\colorbox{red!25}{#1}}
\newcommand{\metrictablesecond}[1]{\colorbox{orange!25}{#1}}
\newcommand{\metrictablethird}[1]{\colorbox{yellow!25}{#1}}

\newcommand{\ctext}[3][RGB]{%
  \begingroup
  \definecolor{hlcolor}{#1}{#2}\sethlcolor{hlcolor}%
  \hl{#3}%
  \endgroup
}

\DeclareMathOperator*{\argmax}{argmax}

\usepackage[pagebackref,breaklinks,colorlinks,allcolors=cvprblue]{hyperref}
\usepackage[table]{xcolor}
\usepackage{multirow}
\usepackage{soul,xcolor}

\title{SplatSuRe: Selective Super-Resolution for Multi-view Consistent \\ 3D Gaussian Splatting}

\author{%
  Pranav Asthana
  \hspace{2em}
  Alex Hanson
  \hspace{2em}
  Allen Tu
  \hspace{2em}
  Tom Goldstein
  \\
  Matthias Zwicker 
  \hspace{2em}
  Amitabh Varshney\\
    [0.35em]
    \textnormal{University of Maryland, College Park} \\
    [0.4em]
   \url{https://splatsure.github.io}  
  \vspace{-8mm}
}

\begin{document}

\twocolumn[{%
    \renewcommand\twocolumn[1][]{#1}%
    \maketitle
    \includegraphics[width=\linewidth]{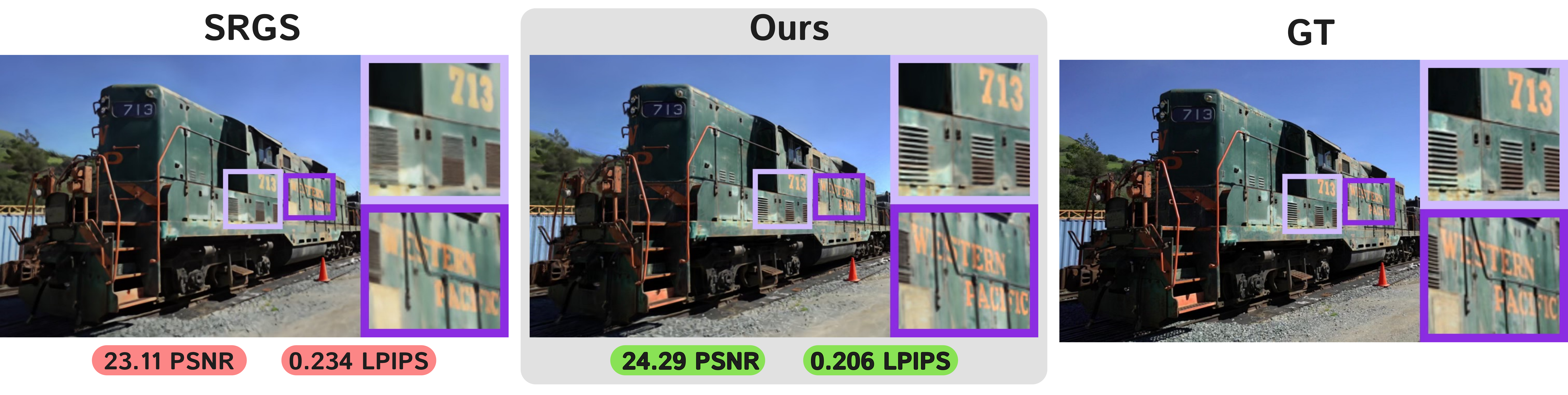}
    \vspace{-8mm}
    \captionsetup{type=figure}
    \caption{\textbf{SplatSuRe} trains a 3D Gaussian Splatting~\cite{kerbl3Dgaussians} model to produce sharp, high-resolution novel views from low-resolution inputs. By selectively leveraging high-frequency cues already present in low-resolution training views and applying super-resolution only where needed, our method delivers greater detail and multi-view consistency than prior approaches without any additional training.}
    \label{fig:teaser}
    \vspace{6mm}
}]

\begin{abstract}
3D Gaussian Splatting (3DGS) enables high-quality novel view synthesis, motivating interest in generating higher-resolution renders than those available during training. A natural strategy is to apply super-resolution (SR) to low-resolution (LR) input views, but independently enhancing each image introduces multi-view inconsistencies, leading to blurry renders. Prior methods attempt to mitigate these inconsistencies through learned neural components, temporally consistent video priors, or joint optimization on LR and SR views, but all uniformly apply SR across every image. In contrast, our key insight is that close-up LR views may contain high-frequency information for regions also captured in more distant views and that we can use the camera pose relative to scene geometry to inform where to add SR content. Building on this insight, we propose SplatSuRe, a method that selectively applies SR content only in undersampled regions lacking high-frequency supervision, yielding sharper and more consistent results. Across Tanks \& Temples, Deep Blending, and Mip-NeRF 360, our approach surpasses baselines in both fidelity and perceptual quality. Notably, our gains are most significant in localized foreground regions where higher detail is desired.

\end{abstract}    
\vspace{-2mm}
\section{Introduction}
\label{sec:intro}

Novel view synthesis aims to render unseen viewpoints given a set of multi-view images. 3D Gaussian Splatting (3DGS)~\cite{kerbl3Dgaussians} enables real-time, photorealistic novel view synthesis by representing scenes as an explicit set of anisotropic Gaussians optimized through differentiable splatting. While 3DGS excels at efficiency and reconstruction quality, its performance is tightly coupled to the resolution of the training images. Models trained on low-resolution (LR) inputs lack access to high-frequency signals present in high-resolution (HR) views, resulting in blurry textures, over-smoothed surfaces, and aliasing artifacts when rendered at higher test-time resolutions~\cite{Yu2024MipSplatting}.

When restricted to only LR views, a natural strategy is to apply super-resolution (SR) to enhance them before fitting a 3D model~\cite{feng2024srgssuperresolution3dgaussian}. However, single-image SR operates independently on each view and frequently introduces view-dependent hallucinated textures~\cite{wang2024exploiting}. When these inconsistent SR predictions are used as direct supervision, the resulting 3D optimization receives conflicting gradients across viewpoints, degrading model quality. Prior methods attempt to mitigate these inconsistencies through learned neural components, temporally consistent video priors, or joint optimization on LR and SR views~\cite{Shen2024SuperGaussian,wan2025s2gaussian,xie2024supergssuperresolution3dgaussian,huang2024assrnerfarbitraryscalesuperresolutionvoxel,yu2024gaussiansr}. However, these methods inject SR content \emph{uniformly} across the image, regardless of whether a region actually benefits from generative detail or is already well-constrained by existing LR observations.

In contrast, the key observation motivating our work is that images of a scene do not sample 3D content uniformly. A low-resolution view captured up close often contains enough high-frequency detail to supervise rendering of more distant views that observe the same region only coarsely. This disparity in multi-view sampling implies that many views already receive high-frequency supervision from closer views, whereas others that do not have any closer views that provide sufficient higher-frequency information would benefit from SR guidance. Applying SR indiscriminately therefore introduces unnecessary inconsistencies in well-resolved regions. 

Based on this observation, we propose \textbf{SplatSuRe}, a selective super-resolution framework for multi-view consistent 3D Gaussian Splatting. Rather than uniformly enhancing all pixels, SplatSuRe identifies 3D regions that lack high-frequency observations and injects SR only where it is beneficial. 
We first compute a \emph{Gaussian fidelity score} that measures how well each Gaussian is sampled across training views, then render per-view \emph{super-resolution region selection} weight maps that highlight undersampled areas while suppressing SR where LR supervision is already reliable. These maps modulate SR supervision during training, allowing the model to exploit generative detail in under-resolved regions while maintaining consistency elsewhere. Through this geometry-aware selective refinement, SplatSuRe produces sharper reconstructions and improved perceptual quality without introducing additional neural components or modifying the underlying 3DGS pipeline.

In summary, we propose the following contributions:
\begin{enumerate}
    \item A per-Gaussian fidelity score that quantifies how well it is resolved across views, leveraging LR geometry to estimate available high-frequency information.
    \item A per-view spatial map of available frequency information computed using the Gaussian fidelity score that modulates SR supervision during optimization. 
    \item A selective SR training framework that jointly optimizes a 3DGS model using LR and SR supervision, injecting generative detail only where needed while preserving multi-view consistency and achieving state-of-the-art results across a diverse range of scenes.
\end{enumerate}
\begin{figure*}[ht!]
    \centering
    \includegraphics[width=\linewidth]{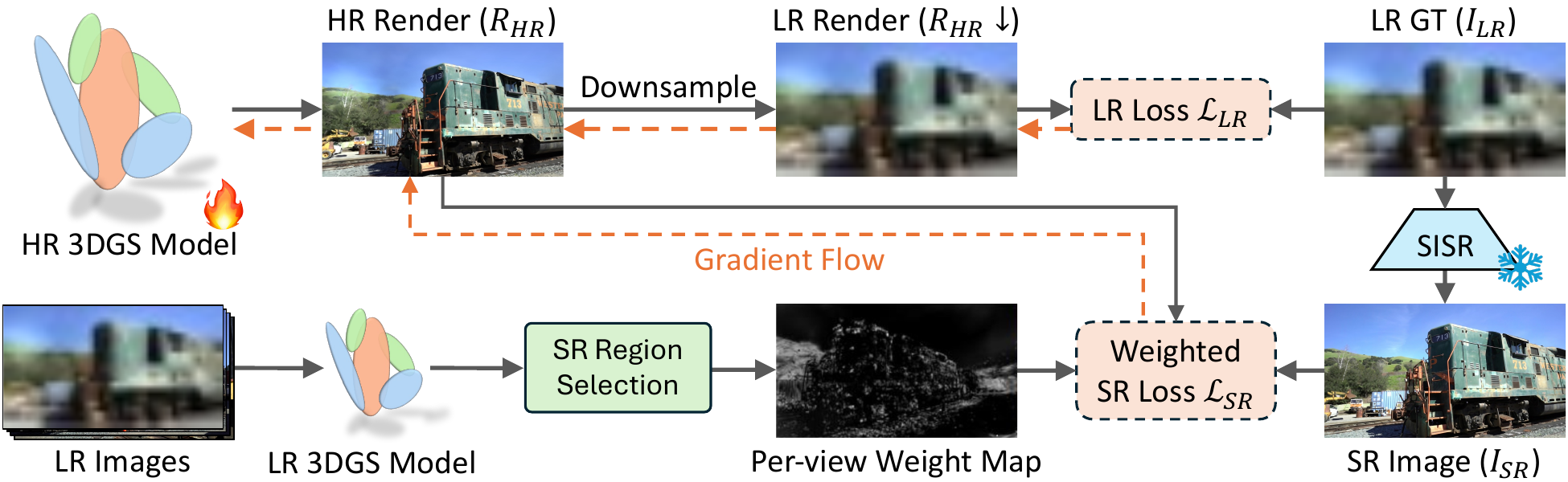}
    \vspace{-6mm}
    \caption{\textbf{Overview of our SplatSuRe framework.} A high-resolution (HR) 3D Gaussian Splatting (3DGS) model is trained using low-resolution (LR) and super-resolution (SR) inputs.
    We first train a 3DGS model on LR inputs to identify undersampled regions and render per-view weight maps that indicate where SR is needed. During training of the HR 3DGS model, the images produced by the frozen single-image super-resolution (SISR) model are spatially weighted by these maps to form the SR loss $\mathcal{L}_{SR}$. A complementary LR loss $\mathcal{L}_{LR}$ compares the downsampled HR render against the original LR ground truth to provide consistent supervision across the entire image.}
    \vspace{-3mm}
    \label{fig:image_pipeline}
\end{figure*}

\section{Related Work}

3D Gaussian Splatting (3DGS)~\cite{kerbl3Dgaussians} enables real-time novel view synthesis by representing scenes as sets of anisotropic Gaussians optimized through differentiable rasterization. While it achieves high rendering efficiency, models trained with low-resolution (LR) images suffer from aliasing artifacts when rendered at higher resolutions, since Gaussians optimized at coarse scales become undersampled. Mip-Splatting~\cite{Yu2024MipSplatting} mitigates this aliasing by applying scale-adaptive 3D and 2D filtering while preserving radiance energy across resolutions, similar in spirit to Mip-NeRF~\cite{barron2022mipnerf360}. While achieving significant improvement over vanilla 3DGS, its rendering quality is still tied to the resolution of training views and blurring can occur at higher resolutions. In contrast, our method uses super-resolution to further inform high-resolution information.

Super-resolution (SR) has long been applied to neural radiance fields~\cite{mildenhall2021nerf} to enhance novel view synthesis quality~\cite{wang2022nerf, 10205402, feng2023zssrtefficientzeroshotsuperresolution, vishen2025advancingsuperresolutionneuralradiance, zheng2025supernerfganuniversal3dconsistentsuperresolution, huang2024assrnerfarbitraryscalesuperresolutionvoxel,roessle2023ganerf}. More recently, SR has been extended to 3D Gaussian Splatting (3DGS). Several variants aim to recover high-resolution detail and improve multi-view consistency~\cite{feng2024srgssuperresolution3dgaussian, Shen2024SuperGaussian,wan2025s2gaussian,xie2024supergssuperresolution3dgaussian} through residual feature learning~\cite{xie2024supergssuperresolution3dgaussian}, uncertainty modeling~\cite{wan2025s2gaussian, xie2024supergssuperresolution3dgaussian}, per-scene refinement~\cite{wan2025s2gaussian} or video super-resolution~\cite{Shen2024SuperGaussian}. Among these, SRGS~\cite{feng2024srgssuperresolution3dgaussian} jointly optimizes Gaussian parameters using both LR ground-truth images and super-resolved views produced by a frozen single-image SR model. However, applying this enhancement uniformly across all regions does not eliminate the effect of inconsistencies introduced by super-resolution. S2Gaussian~\cite{wan2025s2gaussian} focuses on sparse view reconstruction and proposes an inconsistency modeling module trained per-scene to reduce inconsistencies in SR images. SuperGaussian~\cite{Shen2024SuperGaussian} applies a video SR network to frames rendered from a low-resolution 3DGS model, using the resulting temporally consistent sequence for retraining the 3D model. While these approaches improve fidelity, they either rely on additional neural components to enforce consistency or lack spatial adaptivity in the use of SR. In contrast, our method utilizes camera pose information relative to scene geometry to determine undersampled regions, and selectively applies super-resolution in those regions to enhance sharpness while maintaining fidelity.

Diffusion methods offer complementary advances in 3D representation and enhancement tasks. Several works jointly optimize diffusion and 3D parameters for view-consistent generation and super-resolution~\cite{lin2025diffsplat,DiffGS,yu2024gaussiansr,chen2025bridgingdiffusionmodels3d}. GaussianSR~\cite{yu2024gaussiansr} distills information from 2D super-resolution models as a loss for training a 3DGS model. 3DSR~\cite{chen2025bridgingdiffusionmodels3d} uses a 3DGS model to enforce 3D consistency in the diffusion process, fitting it multiple times through denoising diffusion steps. Diffusion priors have also been applied as post-processing to enhance renders from 3DGS models~\cite{liu20243dgs,wu2025difix3d,gsfix3d}. While these methods rely on large pretrained diffusion models to predict images that reduce 3D inconsistencies, we leverage explicit geometric relationships between cameras and scene structure to determine where generative detail is needed, which can then be added via any of these complementary methods.

In addition to image and video-based models, SR can also be performed directly in 3D, circumventing multi-view inconsistencies. Geometric point-based networks~\cite{yu2018pu,Liu_2024_CVPR,fang2024egp3dedgeguidedgeometricpreserving,dinesh20193dpointcloudsuperresolution} locally upscale point clouds to recover fine-grained geometry. However, these approaches typically only upsample geometry, while texture resolution still relies on 2D methods. As before, our method is orthogonal to these techniques and can be combined with them to further improve reconstruction fidelity.

\section{Background}

\subsection{3D Gaussian Splatting}

3D Gaussian Splatting (3DGS)~\cite{kerbl3Dgaussians} models a scene as a set of 3D Gaussians, each represented by a mean position $\boldsymbol{\mu} \in \mathcal{R}^3$, per-axis scale $\boldsymbol{s} \in \mathcal{R}^3$, rotation vector $\boldsymbol{q} \in \mathcal{R}^4$, scalar opacity $o \in \mathcal{R}_+$, and view-dependent color $\boldsymbol{c}$, represented as a base color with spherical harmonic coefficients. Volumetric rendering with alpha blending is used to splat the Gaussians onto the image plane and the resulting color for pixel $\boldsymbol{p}$ is given by:
\begin{align}
    C(\boldsymbol{p}) = \sum_{i=1}^{N} \boldsymbol{c}^i\alpha^i(\boldsymbol{p}) \prod_{j=1}^{i-1} (1-\alpha^j(\boldsymbol{p})),
\end{align}
where $N$ is the number of Gaussians intersecting the pixel ray, 
$\alpha^i(\boldsymbol{p}) = o^ie^{-\frac{1}{2}(\boldsymbol{p} - \boldsymbol{\mu}^i)^T(\boldsymbol{\Sigma}_{{2D}}^i)^{-1}(\boldsymbol{p} - \boldsymbol{\mu}^i)}$ is the contribution of the $i^{th}$ Gaussian, and $\boldsymbol{c}^i$ is its view-dependent color. The 2D covariance matrix $\boldsymbol{\Sigma}_{2D}$ is given by the EWA Splatting approximation~\cite{zwicker2002ewa}. Gaussians are sorted in depth order before splatting to ensure that transmittance is computed correctly.

The model is initialized using Structure from Motion (SfM) on the training views to provide camera parameters and produce a sparse point cloud, serving as the initial Gaussian means. During model training, images are randomly selected from the training set, and a weighted sum of $\mathcal{L}_1$ and $\mathcal{L}_{\text{D-SSIM}}$ losses is used to optimize the Gaussian parameters using gradient descent. Gaussians are split and cloned throughout training to increase scene fidelity and ensure sufficient primitives where needed. We retain the standard 3DGS rendering and densification pipeline, modifying only the supervision losses for our method.

\subsection{Super-Resolution for Gaussian Splatting}
When trained solely on low-resolution (LR) images, 3DGS models lack access to the high-frequency cues present in high-resolution (HR) views, leading to over-smoothed textures and incomplete fine structure. This limitation motivates integrating super-resolution (SR) models into the 3DGS training pipeline. SRGS~\cite{feng2024srgssuperresolution3dgaussian} employs a frozen single-image SR model to generate SR views and jointly optimizes Gaussian parameters against both LR ground truth and SR outputs, supplying explicit high-frequency supervision that LR views alone cannot provide. However, SRGS applies SR \emph{uniformly} across the entire image, even in regions that already receive reliable high-frequency supervision from nearby LR views. Since SR is applied independently to each image, the generated details are not necessarily multi-view consistent -- injecting them everywhere can introduce geometric or texture inconsistencies, leading to averaging effects in the model that render as blurring.

These observations reveal a fundamental challenge: \emph{super-resolution is not uniformly beneficial across the scene}. As illustrated in Figure~\ref{fig:image_disparity}, some regions already receive sufficient high-frequency supervision from closer LR training views, so adding generated detail introduces unnecessary inconsistencies that harm cross-view coherence. Other regions, particularly those that are distant or sparsely observed, are undersampled and require SR to recover missing details. This motivates a selective, geometry-aware strategy that determines where SR should influence optimization. Instead of treating all pixels equally, we aim to exploit the multi-view sampling pattern of each Gaussian to determine which regions are sufficiently constrained by LR observations and which require additional SR guidance.
\section{Method}
\label{sec:method}
Super-resolution (SR) mostly benefits regions that lack reliable high-frequency supervision, motivating a geometry-aware mechanism for deciding where SR should influence 3DGS optimization. Our method identifies these undersampled areas and applies SR only where high-frequency detail is missing, improving sharpness while avoiding unnecessary inconsistencies. To achieve this, we compute a Gaussian fidelity score that measures how well each Gaussian is sampled across training views, then render per-view weight maps that highlight undersampled areas while suppressing SR where LR supervision is already reliable. Incorporating these weight maps into a combined LR–SR objective yields our \textbf{SplatSuRe} framework, shown in Figure~\ref{fig:image_pipeline}, which selectively injects detail where it is beneficial while preserving multi-view consistency elsewhere.

\subsection{Gaussian Fidelity Score}

\begin{figure}[t!]
    \centering
    \includegraphics[width=0.8\linewidth]{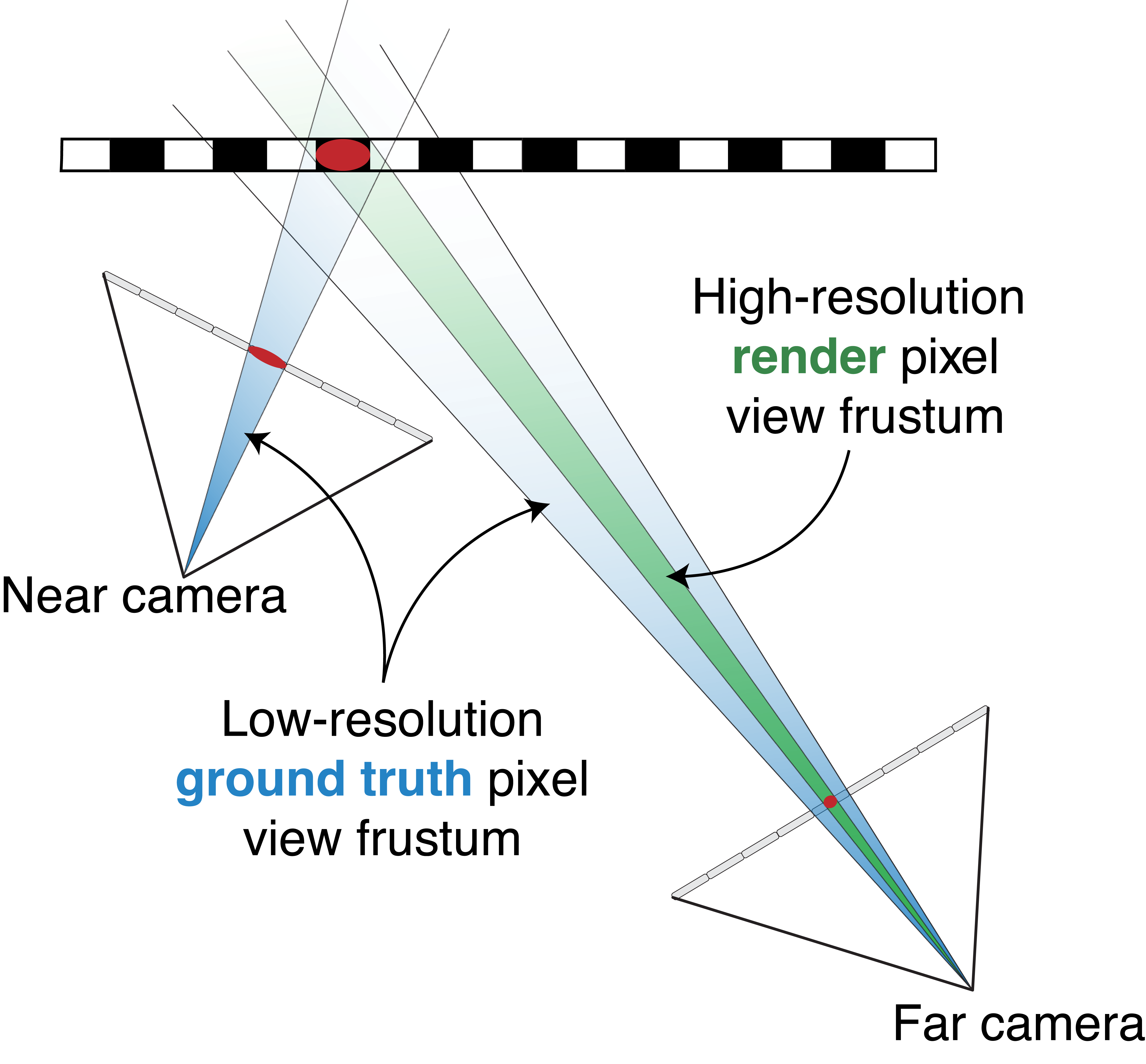}
    \vspace{-2mm}
    \caption{\textbf{Disparity in high-frequency ground truth information across different views.} Low-resolution ground truth from near cameras provides high-resolution information for rendering distant views, reducing the need for additional generated detail in those views. Conversely, super-resolution is needed in views where no other camera provides higher-resolution information.}
    \label{fig:image_disparity}
    \vspace{-2mm}
\end{figure}

Images capturing a scene do not contribute equal amounts of high-frequency information for 3D reconstruction. A low-resolution (LR) view taken at a short distance or with a long focal length can contain more fine detail than a high-resolution (HR) view captured from further away. We leverage this inherent disparity to determine which 3D regions already have adequate high-frequency supervision and which require additional details from super-resolution (SR). Figure~\ref{fig:image_disparity} illustrates how a nearby LR image can provide the high-frequency detail needed to supervise a distant viewpoint.

To measure per-Gaussian relative sampling frequency, we first train a low-resolution 3DGS model using the LR images. This provides stable scene geometry and allows us to compute each Gaussian's screen-space radius in every training view. Following 3DGS, the screen-space radius, measured in pixel units, of Gaussian $\mathcal{G}^i$ is:
\begin{align}
    r^i = 3\sqrt{\text{max}(\lambda^i_1, \lambda^i_2)},
\end{align}
where $\lambda^i_1$ and $\lambda^i_2$ are the eigenvalues of the Gaussian's 2D covariance matrix $\boldsymbol{\Sigma}^i_{2D}$, calculated as:
\begin{align}
    \lambda^i_1, \lambda^i_2 =& \frac{1}{2}\text{tr}(\boldsymbol{\Sigma}^i_{2D}) \pm \sqrt{\text{max}\{0.1, \frac{1}{4}\text{tr}^2(\boldsymbol{\Sigma}^i_{2D})-|\boldsymbol{\Sigma}^i_{2D}|\}},
\end{align}
where tr$(\boldsymbol{\Sigma}^i_{2D})$ denotes the trace of the projected covariance matrix.
For each Gaussian, we then compute the ratio $\rho^i$ between its maximal and minimal radius across all training views $T$ in which it contributes to the rendering:
\begin{equation}
    r^i_{min} = \min_{t\in T}{r^i_t}, \qquad 
    r^i_{max} = \max_{t\in T}{r^i_t},
\end{equation}
\begin{equation}
    \rho^i = r^i_{max}/r^i_{min}.
\end{equation}
We use this ratio as an approximation for the Gaussian's sampling frequency across views. 
A high ratio indicates that the Gaussian is sampled at varying frequencies, meaning some views observe it with high fidelity and can supervise the others. Conversely, a ratio close to one indicates uniform sampling, indicating that this region lacks any higher-frequency observations and requires SR to add generated details. This interpretation also holds for view regions where the Gaussian projects near its maximal radius, since no other views provide higher frequency information.

Note that 3DGS dilates each Gaussian by convolving it with a fixed low-pass Gaussian filter to prevent aliasing and ensure a minimal rendering size:
\begin{equation}
    \boldsymbol{\Sigma}^i_{2D} = \boldsymbol{\Sigma}^i_{2D} + s\boldsymbol{I},
\end{equation}
where $s{=}0.3$ is the amount of dilation.
Since this dilation artificially inflates the radius, especially for distant Gaussians where the blur dominates, we exclude this dilation when computing radii for the ratio.

We then transform the raw ratio $\rho^i$ into a \emph{Gaussian fidelity score} that maps each Gaussian to a weight in $[0,1]$, with lower values indicating a greater need for SR.
The ratio is first offset by a threshold $\tau$ and then mapped into the unit interval using a sigmoid function, giving the per-Gaussian score:
\begin{align}
    \text{score}_{\mathcal{G}^i} = \sigma(\frac{\rho^i-\tau}{k}),
\end{align}
where $\sigma$ is the sigmoid function and $k{=}0.05$ controls the smoothness of the transition from 0 to 1. The \emph{ratio threshold} $\tau$ is a hyper-parameter that depends on the structure of the scene and the consistency of the SR model across views. Scores for Gaussians visible in fewer than three views are set to zero because these regions are poorly constrained. Higher scores correspond to Gaussians that are already well-captured by LR supervision, while lower scores identify regions where SR should be applied more heavily.

\subsection{Super-Resolution Region Selection}

\begin{figure}[ht!]
    \centering
    \includegraphics[width=0.49\linewidth]{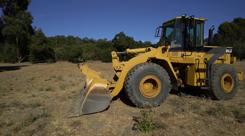}
    \includegraphics[width=0.49\linewidth]{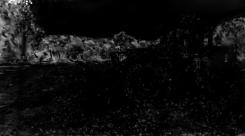}
    \includegraphics[width=0.49\linewidth]{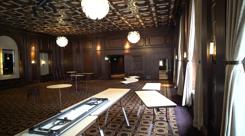}
    \includegraphics[width=0.49\linewidth]{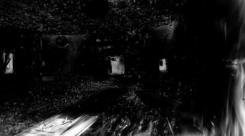}
    \vspace{-2mm}
    \caption{\textbf{Super-resolution weight maps.} Bright regions indicate areas where generative detail is required, while dark regions correspond to areas well-sampled by other low-resolution views. Note that high weights are obtained in regions that are either not sampled closely, such as background trees behind the tractor, or where other views do not provide higher resolution information, such as the foreground table in the ballroom.}
    \label{fig:weight_maps}
    \vspace{-2mm}
\end{figure}

Our Gaussian fidelity score provides a scene-level measure of how well a Gaussian is observed across views, but supervision of high-resolution model updates requires pixel-wise weight maps for each training view. For a given training view $t$, we identify the set of Gaussians whose maximal radius occurs in its rendered view:
\begin{align}
    \mathcal{M}(t) &= \{\mathcal{G}^i\ |\ t = \argmax_{t^{'} \in T}r^i_{t^{'}} \; \forall i \in \{1, \ldots, N\} \},
\end{align}
where $\mathcal{G}^i$ is the $i^{th}$ Gaussian, $r^i_t$ is its screen-space radius, and $N$ is the total number of Gaussians. These are Gaussians that do not receive higher-frequency information from another view. The weight map for training view $t$ is then rendered as:
\begin{align}
    W^{\prime}_t &= (1-\text{Render}(\mathbf{score}_\mathcal{G})) + \text{Render}(\mathbf{1}_{\mathcal{M}(t)}(\mathcal{G})),
\label{eq:eq_weight_map}
\end{align}
where Render$(\cdot)$ denotes splatting the LR model by alpha-blending the specified per-Gaussian values, $\mathcal{G}$ is the set of all Gaussians, $\mathbf{score}_\mathcal{G}$ is the vector of Gaussian fidelity scores, and $\mathbf{1}_{\mathcal{M}(t)}(\mathcal{G})$ is an indicator that is 1 for Gaussians in $\mathcal{M}(t)$ and 0 otherwise. 
The first term in Equation~\ref{eq:eq_weight_map} ensures that undersampled regions with low fidelity scores receive SR, while the second term ensures that areas observed most closely by the current view also receive SR because no alternative view provides higher resolution information. Finally, the weight map is normalized to ensure that the magnitude of the SR loss is consistent across views. 
As illustrated in Figure~\ref{fig:weight_maps}, this per-view weight map is high in regions requiring SR and low in regions already sufficiently supervised by LR views.

\subsection{SplatSuRe Training Objective}

Our super-resolution region selection method determines how much SR should influence each pixel during optimization.
Rather than uniformly applying SR to the entire image during training, we incorporate our weight map $W_t$ to supervise training using two complementary signals: (1) low-resolution ground truth images, and (2) selectively-weighted super-resolved images. 

In each training iteration, the model is rendered at the target high-resolution. The render $R_{HR}$ is downsampled to produce $R_{HR}\downarrow$ and compared with the LR ground truth $I_{LR}$, yielding the LR loss:
\begin{equation}
    \mathcal{L}_{LR} = (1-\lambda)\mathcal{L}_1(R_{HR}\downarrow, I_{LR}) + \lambda\mathcal{L}_{\text{D-SSIM}}(R_{HR}\downarrow, I_{LR}),\label{eq_L_lr}
\end{equation}
where $\mathcal{L}_1$ and $\mathcal{L}_{\text{D-SSIM}}$ are the losses used in 3DGS. The super-resolved image $I_{SR}$ is compared with the HR render $R_{HR}$ using a spatially weighted loss:
\begin{equation}
    \mathcal{L}_{SR} = (1-\lambda)\mathcal{L}_1^W(R_{HR}, I_{SR}) + \lambda\mathcal{L}_{\text{D-SSIM}}^W(R_{HR}, I_{SR}),\label{eq_L_sr}
\end{equation}
where each pixel's contribution is scaled by its weight in $W_t$. High weights amplify SR supervision in undersampled regions, while low weights suppress it where LR views already provide reliable high-frequency information.

The overall objective combines LR ground truth supervision with selectively-weighted SR guidance:
\begin{align}
    \mathcal{L} = (1-\gamma)\mathcal{L}_{LR} + \gamma\mathcal{L}_{SR},\label{eq_L_total}
\end{align}
where $\gamma$ controls the relative contribution of each term. This \textbf{SplatSuRe} formulation produces a high-resolution 3DGS model that leverages SR only where it improves reconstruction quality, avoiding inconsistencies in regions already well-constrained by LR views.
\begin{figure*}
    \centering
    \includegraphics[width=\linewidth]{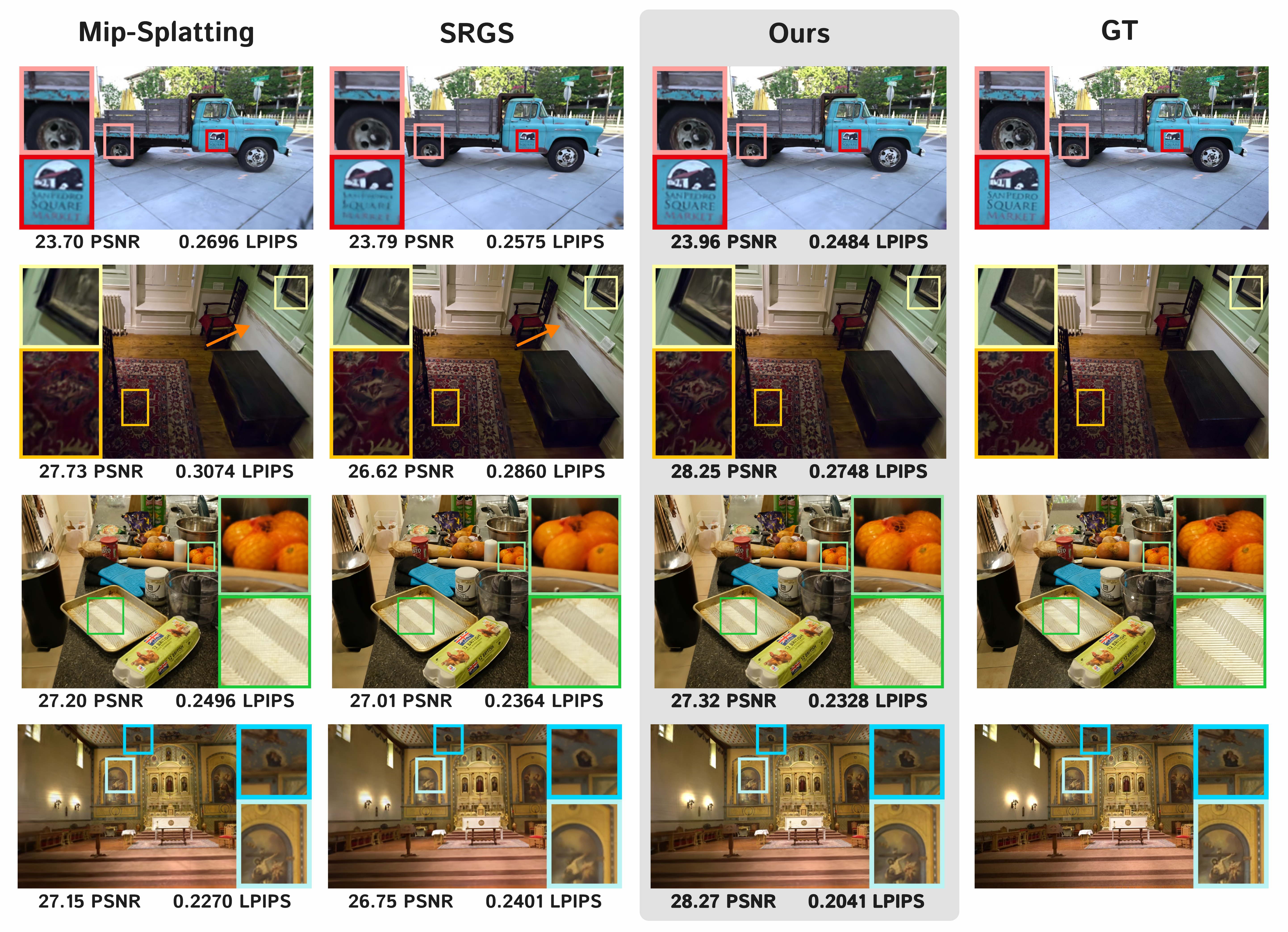}
    \vspace{-8mm}
    \caption{\textbf{Qualitative results on Tanks \& Temples~\cite{Knapitsch2017tandt}, Deep Blending~\cite{DeepBlending2018}, and Mip-NeRF 360~\cite{barron2022mipnerf360}.} Experiments are performed at $4\times$ super-resolution with ratio threshold $\tau{=}1.1$. Compared to Mip-Splatting~\cite{Yu2024MipSplatting} and SRGS~\cite{feng2024srgssuperresolution3dgaussian}, our method produces sharper, more faithful reconstructions that better align with ground truth while maintaining cross-view consistency. It preserves finer details in text (\ctext[RGB]{255, 89, 89}{red box} on truck), high-frequency patterns (\ctext[RGB]{255, 196, 0}{yellow box} on carpet and \ctext[RGB]{53, 180, 53}{green box} on tray) and distant objects observed in other views (\ctext[RGB]{183, 240, 241}{blue box} on church mural). Notably, it reduces Gaussian artifacts (\ctext[RGB]{246, 124, 2}{orange arrow}) observed in other methods. Additional results in Appendix~\ref{appendix:additional_results}.}
    \label{fig:main_results_fig}
    \vspace{-3mm}
\end{figure*}

\section{Experiments}
\label{sec:experiments}

We integrate our \textbf{SplatSuRe} method into the 3D Gaussian Splatting (3DGS) codebase~\cite{kerbl3Dgaussians}, adding modules for true Gaussian radius computation, weight map computation and rendering, and auxiliary loss terms. Experiments are conducted at $4\times$ super-resolution using StableSR~\cite{wang2024exploiting} and using a ratio threshold $\tau{=}1.1$ to generate weight maps. Our choice of ratio threshold and SR model are ablated in Sections~\ref{sec:ratio_ablation} and \ref{sec:abl:sr_model}, respectively. We use $\lambda{=}0.2$ and $\gamma{=}0.4$ for our losses defined in Equations~\ref{eq_L_lr}-\ref{eq_L_total}.

\vspace{-4mm}
\paragraph{Datasets.}
We evaluate our method on three real-world datasets that provide a diverse mix of indoor and outdoor environments, scenes with both circular and elongated central objects, and camera paths ranging from smooth trajectories to irregular captures. \textbf{Tanks \& Temples}~\cite{Knapitsch2017tandt} includes 21 real-world scenes of varying scales, of which we use 19 -- two are excluded because COLMAP fails. Images are downsampled to $240\times135$ and upsampled $4\times$ to $960\times540$. Each scene contains roughly 150-500 images, with every eighth image used for testing and the remainder for training. \textbf{Mip-NeRF 360}~\cite{barron2022mipnerf360} consists of nine real-world scenes -- five outdoor and four indoor -- with about 250–300 images each at approximately 4K$\times$3K resolution. We downsample these images by $8\times$ ($\sim$500$\times$375) and upsample by $4\times$ to half the native resolution ($\sim$2K$\times$1.5K), again reserving every eighth image for testing. Finally, we use two scenes from \textbf{Deep Blending}~\cite{DeepBlending2018}, which provides a collection of forward-facing indoor scenes captured under diverse lighting conditions. Each scene contains roughly 250 images at $\sim$1K$\times$1K resolution, which we downsample by $4\times$ for
training and evaluate on the original full-resolution renders.

\begin{table*}[t]
\centering
\caption{\textbf{Quantitative results on Tanks \& Temples~\cite{Knapitsch2017tandt}.} Experiments are performed at \textbf{$4\times$} super-resolution using ratio threshold $\tau{=}1.1$. The \metrictablebest{best}, \metrictablesecond{second best} and \metrictablethird{third best} entries are highlighted. Our SplatSuRe method achieves the strongest results across most metrics.}
\resizebox{\textwidth}{!}{
\begin{tabular}{l|cccccccccccc}
\toprule
\multirow{2}{*}{\textbf{Method}} & \multicolumn{8}{c}{\textbf{Tanks \& Temples~\cite{Knapitsch2017tandt}}} \\
& SSIM~$\uparrow$ & PSNR~$\uparrow$ & LPIPS~$\downarrow$ & FID~$\downarrow$ & CMMD~$\downarrow$ & DreamSim~$\downarrow$ & MUSIQ~$\uparrow$ & NIQE~$\downarrow$\\
\midrule
3DGS (LR)~\cite{kerbl3Dgaussians} & 
0.669 & 19.41 & 0.350  & 71.58 & 2.013 & 0.0895 & \metrictablesecond{57.776} & \metrictablebest{3.412}\\
3DGS~\cite{kerbl3Dgaussians} + StableSR~\cite{wang2024exploiting} & 0.751  & 22.47 & \metrictablethird{0.300} & 59.29 & \metrictablethird{1.123} & 0.0667 & \metrictablethird{56.748} & 4.945\\
Mip-Splatting~\cite{Yu2024MipSplatting} & \metrictablethird{0.767} & \metrictablethird{23.10} & 0.303 & \metrictablethird{52.46} & 1.137 & \metrictablethird{0.0597} & 46.571 & 5.043\\
SRGS~\cite{feng2024srgssuperresolution3dgaussian} + StableSR~\cite{wang2024exploiting} & \metrictablesecond{0.771} & \metrictablesecond{23.32} & \metrictablesecond{0.286} & \metrictablesecond{49.11} & \metrictablesecond{1.048} & \metrictablesecond{0.0535} & 55.209 & \metrictablethird{4.633} \\
Ours + StableSR~\cite{wang2024exploiting} & \metrictablebest{0.784} & \metrictablebest{23.81} & \metrictablebest{0.272} & \metrictablebest{37.72} & \metrictablebest{1.040} & \metrictablebest{0.0413} & \metrictablebest{58.332} & \metrictablesecond{3.928}\\
\bottomrule
\end{tabular}
}
\label{tab:tandt}
\vspace{-1mm}
\end{table*}

\begin{table*}[t]
\centering
\caption{\textbf{Quantitative results on Deep Blending~\cite{DeepBlending2018} and Mip-NeRF 360~\cite{barron2022mipnerf360}.} Our SplatSuRe method achieves the strongest results across all metrics on Deep Blending and outperforms SRGS~\cite{feng2024srgssuperresolution3dgaussian} on Mip-NeRF 360. Appendix~\ref{sec:app:8x_sr} and \ref{appendix:per_scene_analysis} present $8\times$ SR and per-scene results.}
\resizebox{\textwidth}{!}{
\begin{tabular}{l|ccccc|ccccc}
\toprule
\multirow{2}{*}{\textbf{Method}} & \multicolumn{5}{c|}{\textbf{Deep Blending~\cite{DeepBlending2018}}} & \multicolumn{5}{c}{\textbf{Mip-NeRF 360~\cite{barron2022mipnerf360}}}\\
&SSIM~$\uparrow$ & PSNR~$\uparrow$ & LPIPS~$\downarrow$ & CMMD~$\downarrow$&DreamSim~$\downarrow$&SSIM~$\uparrow$ & PSNR~$\uparrow$ & LPIPS~$\downarrow$ & CMMD~$\downarrow$&DreamSim~$\downarrow$\\
\midrule
3DGS (LR)~\cite{kerbl3Dgaussians} & 0.836 & 26.72 & 0.335 & 0.868 & 0.0574 & 0.635 & 20.67 & 0.384 & \metrictablethird{0.645} & 0.0529 \\
3DGS~\cite{kerbl3Dgaussians} + StableSR~\cite{wang2024exploiting} & 0.845 & 27.16 & \metrictablethird{0.325} & \metrictablesecond{0.630} & 0.0505 & 0.699 & 24.28 & 0.351 & 0.725 & 0.0387 \\
Mip-Splatting~\cite{Yu2024MipSplatting} & \metrictablesecond{0.865} & \metrictablesecond{28.43} & 0.327 & \metrictablethird{0.690} & \metrictablesecond{0.0398} & \metrictablebest{0.759} & \metrictablebest{26.48} & \metrictablebest{0.292} &  \metrictablebest{0.183} & \metrictablebest{0.0132}\\
SRGS~\cite{feng2024srgssuperresolution3dgaussian} + StableSR~\cite{wang2024exploiting} & \metrictablethird{0.861} & \metrictablethird{28.23} & \metrictablesecond{0.317} & \metrictablesecond{0.630} & \metrictablethird{0.0409} & \metrictablethird{0.734} & \metrictablethird{25.92} & \metrictablesecond{0.317} & \metrictablesecond{0.339} & \metrictablethird{0.0194} \\
Ours + StableSR~\cite{wang2024exploiting} & \metrictablebest{0.872} & \metrictablebest{29.01} & \metrictablebest{0.306} & \metrictablebest{0.496} & \metrictablebest{0.0330} & \metrictablesecond{0.740} & \metrictablesecond{26.34} & \metrictablethird{0.323} & \metrictablesecond{0.339} & \metrictablesecond{0.0179} \\
\bottomrule
\end{tabular}
}
\label{tab:deepblending_mipnerf}
\vspace{-3mm}
\end{table*}

\vspace{-4mm}
\paragraph{Baselines.} We compare SplatSuRe with four representative baselines using their official implementations on our evaluation datasets. \textbf{3DGS (LR)}~\cite{kerbl3Dgaussians} is trained on LR inputs and is the primary baseline. \textbf{3DGS + StableSR}~\cite{wang2024exploiting} applies a single-image SR model to the low-resolution training data and fits a 3DGS model on the super-resolved images. \textbf{Mip-Splatting}~\cite{Yu2024MipSplatting} is trained at low-resolution and mitigates aliasing through 3D and 2D multi-scale filtering for high-resolution rendering. Finally, \textbf{SRGS}~\cite{feng2024srgssuperresolution3dgaussian} optimizes Gaussian parameters using both low-resolution and super-resolved images from a frozen SR model, without additional pretraining or scene-specific fine-tuning. 

\vspace{-4mm}
\paragraph{Metrics.} We evaluate both reconstruction fidelity and perceptual realism using eight complementary metrics. For reference-based assessment, we report SSIM, PSNR, and LPIPS~\cite{zhang2018perceptual}, which measure pixel and feature-level agreement with ground-truth high-resolution images. To capture distributional and semantic perceptual quality, we include FID~\cite{heusel2017gans}, CMMD~\cite{jayasumana2024rethinkingfidbetterevaluation}, and DreamSim~\cite{fu2023dreamsim}, which compare the feature distributions or embeddings of generated and real images in pretrained perceptual spaces. Finally, we evaluate no-reference perceptual quality using MUSIQ~\cite{ke2021musiqmultiscaleimagequality} and NIQE~\cite{6353522}, which estimate realism and naturalness without ground-truth supervision. While PSNR and SSIM directly measure pixel fidelity at the native resolution, the perceptual and distributional metrics internally downsample or resize images before feature extraction, reducing sensitivity to high-frequency details. As a result, improvements in fine-scale sharpness -- central to super-resolution methods -- may be underrepresented by these metrics, motivating our use of a broad set of complementary evaluations.
\section{Results}
\label{sec:results}

\paragraph{Qualitative Results.} Figures~\ref{fig:teaser} and~\ref{fig:main_results_fig} present representative renderings from our method. Our approach produces consistently sharper reconstructions while preserving smoothness in uniformly textured regions. In contrast, Mip-Splatting yields overly blurred results because it is trained only on LR inputs and its 3D anti-aliasing filter excessively blurs all Gaussians. SRGS recovers higher-frequency content in some areas but remains constrained by the underlying SR images -- regions where the SR output is consistent appear sharp, whereas areas with view-inconsistent SR predictions become noticeably blurred or distorted.

\vspace{-4mm}
\paragraph{Quantitative Results.} 
Table~\ref{tab:tandt} presents results on Tanks \& Temples~\cite{Knapitsch2017tandt}. Our method achieves the highest image quality across nearly all metrics, demonstrating both its effectiveness and robustness. Only 3DGS attains a better NIQE score. Similarly, SRGS ranks second across most reference and perceptual metrics but lags behind 3DGS in MUSIQ and NIQE. This trend arises because the aliased, noisy renderings of low-resolution 3DGS coincidentally resemble the natural image statistics favored by no-reference quality metrics, whereas the smoother and more coherent outputs of Mip-Splatting and SR-based methods appear less “natural” under such measures. In contrast, our approach delivers substantially higher perceptual fidelity and cross-view consistency, producing sharper and more realistic high-resolution reconstructions that align closely with both human perception and ground-truth images.

Table~\ref{tab:deepblending_mipnerf} reports results on Deep Blending~\cite{DeepBlending2018} and Mip-NeRF 360~\cite{barron2022mipnerf360}. Our method achieves the strongest performance across all metrics on Deep Blending. On Mip-NeRF 360, it outperforms SRGS on every metric except LPIPS; however, both SR-based approaches are surpassed by Mip-Splatting. This arises from the characteristics of Mip-NeRF 360, where smooth, circular camera trajectories provide dense multi-view coverage and minimal undersampling. In these well-sampled settings, Mip-Splatting’s multi-scale anti-aliasing filters effectively preserve radiance energy and suppress aliasing, whereas SR-based methods may introduce slight inconsistencies when enhancing already well-resolved regions. Additionally, the LR images in Mip-NeRF 360 are roughly twice as large as those in the other datasets, preserving most high-frequency details and leaving little room for SR improvement. Overall, SplatSuRe achieves the highest performance on Tanks \& Temples and Deep Blending and outperforms SRGS on Mip-NeRF 360. Results for $8\times$ SR, additional visualizations, and per-scene metrics for Tables~\ref{tab:tandt} and \ref{tab:deepblending_mipnerf} can be found in Appendix~\ref{sec:app:8x_sr}, \ref{appendix:additional_results}, and \ref{sec:app:scene_metrics}. Appendix~\ref{appendix:unified_pipeline} presents a unified training pipeline that merges the LR initialization and SR refinement stages to achieve comparable performance with the same training budget as single-stage baselines.

\begin{figure}
    \centering
    \includegraphics[width=\linewidth]{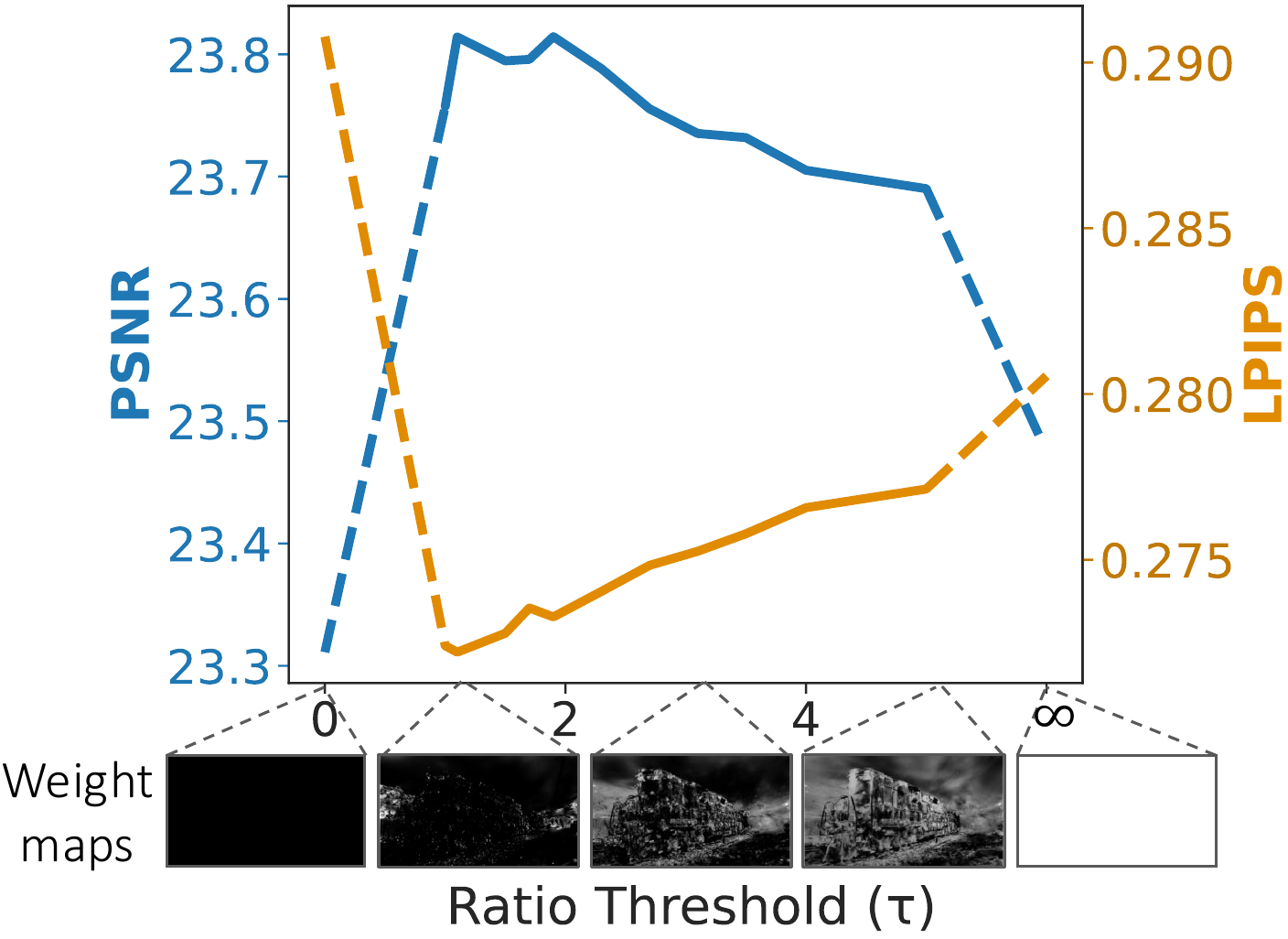}
    \vspace{-7mm}
    \caption{\textbf{Effect of ratio threshold on Tanks \& Temples~\cite{Knapitsch2017tandt}.}
    Weight maps, where bright regions indicate higher SR influence, are shown below the corresponding ratio thresholds. $\tau{=}0$ and $\tau{=}\infty$ correspond to zero and full use of super-resolution. SR is initially helpful in improving rendering quality, but excessive use worsens results. The effect of ratio threshold on different scenes is analyzed in Appendix~\ref{appendix:per_scene_analysis}.} 
    \label{fig:ratio_threshold}
    \vspace{-4mm}
\end{figure}
\section{Ablations}

\subsection{Ablation on Ratio Threshold}
\label{sec:ratio_ablation}

Scenes with different geometries and scales exhibit varying distributions of high and low-frequency content across training views. To account for this variability, we examine the effect of the ratio threshold used to determine where SR is applied. As shown in Figure~\ref{fig:ratio_threshold}, introducing a small amount of SR initially improves both PSNR and LPIPS by recovering fine details in undersampled regions. However, applying SR too aggressively leads to inconsistencies across views, degrading overall performance. This trend is most pronounced in scenes with varied camera-object distances, where sampling density differs significantly across views. Based on this analysis, we select ratio threshold $\tau{=}1.1$, which provides the best trade-off between sharpness and consistency across scenes, in our main experiments. See Appendix~\ref{appendix:per_scene_analysis} for a detailed analysis of different scenes.

\subsection{Ablation on Super-Resolution Model}
\label{sec:abl:sr_model}

We evaluate our method using two single-image super-resolution (SISR) models: SwinIR~\cite{liang2021swinir} and StableSR~\cite{wang2024exploiting}. SwinIR is optimized for high fidelity under pixel-based metrics such as PSNR, producing accurate yet often over-smoothed reconstructions. In contrast, StableSR incorporates diffusion-based generative priors to synthesize sharper, more detailed, and perceptually realistic images, sometimes at the expense of lower PSNR. As shown in Table~\ref{tab:sisr_metrics}, our approach improves performance across both models, demonstrating that it is agnostic to the choice of SISR backbone. Notably, our gains are larger with StableSR, as its higher perceptual quality comes at the cost of multi-view inconsistencies. Across both methods, StableSR produces better perceptual quality metrics, motivating our use of StableSR for the main experiments where perceptual realism is prioritized alongside reconstruction fidelity. 

\begin{table}[t]
\centering
\caption{\textbf{Comparison of SwinIR~\cite{liang2021swinir} and StableSR~\cite{wang2024exploiting} on Tanks \& Temples~\cite{Knapitsch2017tandt}.} Experiments are performed at \textbf{$4\times$} super-resolution using ratio threshold $\tau{=}1.1$. Our method outperforms SRGS with either model. While SwinIR achieves higher PSNR due to its conservative reconstruction, we choose StableSR for our main experiments for its superior perceptual quality.}
\vspace{-2mm}
\resizebox{\columnwidth}{!}{
\begin{tabular}{l|cc|cc}
\toprule
\multirow{2}{*}{\textbf{Method}} & \multicolumn{2}{c|}{\textbf{SwinIR~\cite{liang2021swinir}}} & \multicolumn{2}{c}{\textbf{StableSR~\cite{wang2024exploiting}}}\\
& PSNR~$\uparrow$ & CMMD~$\downarrow$ & PSNR~$\uparrow$ & CMMD~$\downarrow$\\
\midrule
SRGS~\cite{feng2024srgssuperresolution3dgaussian} & 23.84 & 1.137 & 23.32 & 1.048\\
Ours & \textbf{24.06} & \textbf{1.135} & \textbf{23.81} & \textbf{1.043}\\
\bottomrule
\end{tabular}
}
\label{tab:sisr_metrics}
\vspace{-3mm}
\end{table}
\section{Limitations and Future Work}

Our method reduces multi-view inconsistencies by suppressing SR in regions already well-captured by LR views, but this conservative strategy may miss useful SR refinements, such as stable detail along high-contrast boundaries. Selectively applying SR in these regions could further improve reconstruction quality. In addition, our framework operates at a single upsampling level.  Extending it to a multi-scale formulation could provide finer control over SR integration and improve sharpness. Finally, while SplatSuRe produces multi-view consistent 3D models faithful to ground truth, it could further benefit from advances in SR that reduce multi-view inconsistency in generative outputs. 

\section{Conclusion}

We introduced \textbf{SplatSuRe}, a selective super-resolution framework for 3D Gaussian Splatting that applies SR only where high-frequency information is missing. By leveraging scene geometry and camera pose to estimate per-Gaussian sampling fidelity, SplatSuRe identifies undersampled regions and modulates SR supervision through per-view weight maps that highlight where generative detail is truly needed. 
This geometry-aware strategy injects enhanced detail and sharpness where required while preserving multi-view consistency, achieving state-of-the-art results across a diverse range of scenes.
\clearpage
\section*{Acknowledgments}
\label{sec:acknowledgments}

This work was made possible by NSF Grants 21-37229 and 22-35050, DARPA TIAMAT, and the NSF TRAILS Institute (2229885). This research is based upon work supported by the Office of the Director of National Intelligence (ODNI), Intelligence Advanced Research Projects Activity (IARPA), via IARPA R\&D Contract No. 140D0423C0076. The views and conclusions contained herein are those of the authors and should not be interpreted as necessarily representing the official policies or endorsements, either expressed or implied, of the ODNI, IARPA, or the U.S. Government. The U.S. Government is authorized to reproduce and distribute reprints for Governmental purposes notwithstanding any copyright annotation thereon.  Additional support was provided by Coefficient Giving.
{
    \small
    \bibliographystyle{ieeenat_fullname}
    \bibliography{main}

@String(CVPR= {IEEE Conf. Comput. Vis. Pattern Recog.})

@String(ECCV= {Eur. Conf. Comput. Vis.})

@String(ICIP = {IEEE Int. Conf. Image Process.})

@String(ICLR = {Int. Conf. Learn. Represent.})

@String(CVPR  = {CVPR})

@String(ECCV  = {ECCV})

@String(ICIP  = {ICIP})

@String(ICLR  = {ICLR})

@inproceedings{chen2025bridgingdiffusionmodels3d,
  title={Bridging Diffusion Models and 3D Representations: A 3D Consistent Super-Resolution Framework},
  author={Chen, Yi-Ting and Liao, Ting-Hsuan and Guo, Pengsheng and Schwing, Alexander and Huang, Jia-Bin},
  booktitle={Proceedings of the IEEE/CVF International Conference on Computer Vision},
  pages={13481--13490},
  year={2025}
}

@inproceedings{wan2025s2gaussian,
  title={S2Gaussian: Sparse-View Super-Resolution 3D Gaussian Splatting},
  author={Wan, Yecong and Shao, Mingwen and Cheng, Yuanshuo and Zuo, Wangmeng},
  booktitle={Proceedings of the Computer Vision and Pattern Recognition Conference},
  pages={711--721},
  year={2025}
}

@InProceedings{Yu2024MipSplatting,
    author    = {Yu, Zehao and Chen, Anpei and Huang, Binbin and Sattler, Torsten and Geiger, Andreas},
    title     = {Mip-Splatting: Alias-free 3D Gaussian Splatting},
    booktitle = {Proceedings of the IEEE/CVF Conference on Computer Vision and Pattern Recognition (CVPR)},
    month     = {June},
    year      = {2024},
    pages     = {19447-19456}
}

@inproceedings{Shen2024SuperGaussian,
  title = {SuperGaussian: Repurposing Video Models for 3D Super Resolution},
  author = {Shen, Yuan and Ceylan, Duygu and Guerrero, Paul and Xu, Zexiang and Mitra, {Niloy J.} and Wang, Shenlong and Fr{\"u}hst{\"u}ck, Anna},
  booktitle = {European Conference on Computer Vision (ECCV)},
  year = {2024},
}

@InProceedings{wu2025difix3d,
    author    = {Wu, Jay Zhangjie and Zhang, Yuxuan and Turki, Haithem and Ren, Xuanchi and Gao, Jun and Shou, Mike Zheng and Fidler, Sanja and Gojcic, Zan and Ling, Huan},
    title     = {DIFIX3D+: Improving 3D Reconstructions with Single-Step Diffusion Models},
    booktitle = {Proceedings of the IEEE/CVF Conference on Computer Vision and Pattern Recognition (CVPR)},
    month     = {June},
    year      = {2025},
    pages     = {26024-26035}
}

@article{heusel2017gans,
  title={Gans trained by a two time-scale update rule converge to a local nash equilibrium},
  author={Heusel, Martin and Ramsauer, Hubert and Unterthiner, Thomas and Nessler, Bernhard and Hochreiter, Sepp},
  journal={Advances in neural information processing systems},
  volume={30},
  year={2017}
}

@Article{kerbl3Dgaussians,
      author       = {Kerbl, Bernhard and Kopanas, Georgios and Leimk{\"u}hler, Thomas and Drettakis, George},
      title        = {3D Gaussian Splatting for Real-Time Radiance Field Rendering},
      journal      = {ACM Transactions on Graphics},
      number       = {4},
      volume       = {42},
      month        = {July},
      year         = {2023},
      url          = {https://repo-sam.inria.fr/fungraph/3d-gaussian-splatting/}
}

@misc{feng2024srgssuperresolution3dgaussian,
      title={SRGS: Super-Resolution 3D Gaussian Splatting}, 
      author={Xiang Feng and Yongbo He and Yubo Wang and Yan Yang and Wen Li and Yifei Chen and Zhenzhong Kuang and Jiajun ding and Jianping Fan and Yu Jun},
      year={2024},
      eprint={2404.10318},
      archivePrefix={arXiv},
      primaryClass={cs.CV},
      url={https://arxiv.org/abs/2404.10318}, 
}

@inproceedings{liu20243dgs,
  title = {3DGS-Enhancer: Enhancing Unbounded 3D Gaussian Splatting with View-Consistent 2D Diffusion Priors},
  author = {Liu, Xi and Zhou, Chaoyi and Huang, Siyu},
  booktitle = {Advances in Neural Information Processing Systems (NeurIPS)},
  year = {2024}
}

@article{mildenhall2021nerf,
  title={Nerf: Representing scenes as neural radiance fields for view synthesis},
  author={Mildenhall, Ben and Srinivasan, Pratul P and Tancik, Matthew and Barron, Jonathan T and Ramamoorthi, Ravi and Ng, Ren},
  journal={Communications of the ACM},
  volume={65},
  number={1},
  pages={99--106},
  year={2021},
  publisher={ACM New York, NY, USA}
}

@inproceedings{wang2022nerf,
  title={NeRF-SR: High-Quality Neural Radiance Fields using Supersampling},
  author={Wang, Chen and Wu, Xian and Guo, Yuan-Chen and Zhang, Song-Hai and Tai, Yu-Wing and Hu, Shi-Min},
  booktitle={Proceedings of the 30th ACM International Conference on Multimedia},
  pages={6445--6454},
  year={2022}
}

@INPROCEEDINGS{10205402,
  author={Huang, Xudong and Li, Wei and Hu, Jie and Chen, Hanting and Wang, Yunhe},
  booktitle={2023 IEEE/CVF Conference on Computer Vision and Pattern Recognition (CVPR)}, 
  title={RefSR-NeRF: Towards High Fidelity and Super Resolution View Synthesis}, 
  year={2023},
  volume={},
  number={},
  pages={8244-8253},
  doi={10.1109/CVPR52729.2023.00797}}

@article{feng2023zssrtefficientzeroshotsuperresolution,
  title={ZS-SRT: An efficient zero-shot super-resolution training method for Neural Radiance Fields},
  author={Feng, Xiang and He, Yongbo and Wang, Yubo and Wang, Chengkai and Kuang, Zhenzhong and Ding, Jiajun and Qin, Feiwei and Yu, Jun and Fan, Jianping},
  journal={Neurocomputing},
  volume={590},
  pages={127714},
  year={2024},
  publisher={Elsevier}
}

@misc{zheng2025supernerfganuniversal3dconsistentsuperresolution,
      title={SuperNeRF-GAN: A Universal 3D-Consistent Super-Resolution Framework for Efficient and Enhanced 3D-Aware Image Synthesis}, 
      author={Peng Zheng and Linzhi Huang and Yizhou Yu and Yi Chang and Yilin Wang and Rui Ma},
      year={2025},
      eprint={2501.06770},
      archivePrefix={arXiv},
      primaryClass={cs.CV},
      url={https://arxiv.org/abs/2501.06770}, 
}

@misc{huang2024assrnerfarbitraryscalesuperresolutionvoxel,
      title={ASSR-NeRF: Arbitrary-Scale Super-Resolution on Voxel Grid for High-Quality Radiance Fields Reconstruction}, 
      author={Ding-Jiun Huang and Zi-Ting Chou and Yu-Chiang Frank Wang and Cheng Sun},
      year={2024},
      eprint={2406.20066},
      archivePrefix={arXiv},
      primaryClass={cs.CV},
      url={https://arxiv.org/abs/2406.20066}, 
}

@inproceedings{vishen2025advancingsuperresolutionneuralradiance,
  title={Advancing super-resolution in neural radiance fields via variational diffusion strategies},
  author={Vishen, Shrey and Sarabu, Jatin and Kumar, Saurav and Bharathulwar, Chinmay and Lakshmanan, Rithwick and Srinivas, Vishnu},
  booktitle={Proceedings of the Winter Conference on Applications of Computer Vision},
  pages={336--343},
  year={2025}
}

@misc{xie2024supergssuperresolution3dgaussian,
      title={SuperGS: Super-Resolution 3D Gaussian Splatting Enhanced by Variational Residual Features and Uncertainty-Augmented Learning}, 
      author={Shiyun Xie and Zhiru Wang and Xu Wang and Yinghao Zhu and Chengwei Pan and Xiwang Dong},
      year={2024},
      eprint={2410.02571},
      archivePrefix={arXiv},
      primaryClass={cs.CV},
      url={https://arxiv.org/abs/2410.02571}, 
}

@inproceedings{lin2025diffsplat,
  title={DiffSplat: Repurposing Image Diffusion Models for Scalable 3D Gaussian Splat Generation},
  author={Lin, Chenguo and Pan, Panwang and Yang, Bangbang and Li, Zeming and Mu, Yadong},
  booktitle={International Conference on Learning Representations (ICLR)},
  year={2025}
}

@inproceedings{DiffGS,
      title = {DiffGS: Functional Gaussian Splatting Diffusion},
      author = {Zhou, Junsheng and Zhang, Weiqi and Liu, Yu-Shen},
      booktitle = {Advances in Neural Information Processing Systems (NeurIPS)},
      year = {2024}
  }

@article{gsfix3d,
title={GSFix3D: Diffusion-Guided Repair of Novel Views in Gaussian Splatting}, 
author={Jiaxin Wei and Stefan Leutenegger and Simon Schaefer},
year={2025},
eprint={2508.14717},
archivePrefix={arXiv},
primaryClass={cs.CV},
url={https://arxiv.org/abs/2508.14717},
}

@article{yu2024gaussiansr,
  title={GaussianSR: 3D Gaussian Super-Resolution with 2D Diffusion Priors},
  author={Yu, Xiqian and Zhu, Hanxin and He, Tianyu and Chen, Zhibo},
  journal={arXiv preprint arXiv:2406.10111},
  year={2024}
}

@article{Knapitsch2017tandt,
  title={Tanks and temples: Benchmarking large-scale scene reconstruction},
  author={Knapitsch, Arno and Park, Jaesik and Zhou, Qian-Yi and Koltun, Vladlen},
  journal={ACM Transactions on Graphics},
  volume={36},
  number={4},
  pages={1--13},
  year={2017},
  publisher={ACM New York, NY, USA}
}

@inproceedings{barron2022mipnerf360,
  title={Mip-nerf 360: Unbounded anti-aliased neural radiance fields},
  author={Barron, Jonathan T and Mildenhall, Ben and Verbin, Dor and Srinivasan, Pratul P and Hedman, Peter},
  booktitle={Proceedings of the IEEE/CVF Conference on Computer Vision and Pattern Recognition},
  pages={5470--5479},
  year={2022}
}

@article{DeepBlending2018,
  title={Deep blending for free-viewpoint image-based rendering},
  author={Hedman, Peter and Philip, Julien and Price, True and Frahm, Jan-Michael and Drettakis, George and Brostow, Gabriel},
  journal={ACM Transactions on Graphics},
  volume={37},
  number={6},
  pages={1--15},
  year={2018},
  publisher={ACM New York, NY, USA}
}

@inproceedings{yu2018pu,
     title={PU-Net: Point Cloud Upsampling Network},
     author={Yu, Lequan and Li, Xianzhi and Fu, Chi-Wing and Cohen-Or, Daniel and Heng, Pheng-Ann},
     booktitle = {Proceedings of IEEE Conference on Computer Vision and Pattern Recognition (CVPR)},
     year = {2018}
}

@misc{fang2024egp3dedgeguidedgeometricpreserving,
      title={EGP3D: Edge-guided Geometric Preserving 3D Point Cloud Super-resolution for RGB-D camera}, 
      author={Zheng Fang and Ke Ye and Yaofang Liu and Gongzhe Li and Xianhong Zhao and Jialong Li and Ruxin Wang and Yuchen Zhang and Xiangyang Ji and Qilin Sun},
      year={2024},
      eprint={2412.11680},
      archivePrefix={arXiv},
      primaryClass={cs.CV},
      url={https://arxiv.org/abs/2412.11680}, 
}

@InProceedings{Liu_2024_CVPR,
    author    = {Liu, Yanzhe and Chen, Rong and Li, Yushi and Li, Yixi and Tan, Xuehou},
    title     = {SPU-PMD: Self-Supervised Point Cloud Upsampling via Progressive Mesh Deformation},
    booktitle = {Proceedings of the IEEE/CVF Conference on Computer Vision and Pattern Recognition (CVPR)},
    month     = {June},
    year      = {2024},
    pages     = {5188-5197}
}

@inproceedings{dinesh20193dpointcloudsuperresolution,
  title={3D point cloud super-resolution via graph total variation on surface normals},
  author={Dinesh, Chinthaka and Cheung, Gene and Baji{\'c}, Ivan V},
  booktitle={2019 IEEE international conference on image processing (ICIP)},
  pages={4390--4394},
  year={2019},
  organization={IEEE}
}

@inproceedings{zhang2018perceptual,
  title={The unreasonable effectiveness of deep features as a perceptual metric},
  author={Zhang, Richard and Isola, Phillip and Efros, Alexei A and Shechtman, Eli and Wang, Oliver},
  booktitle={Proceedings of the IEEE conference on computer vision and pattern recognition},
  pages={586--595},
  year={2018}
}

@inproceedings{jayasumana2024rethinkingfidbetterevaluation,
  title={Rethinking fid: Towards a better evaluation metric for image generation},
  author={Jayasumana, Sadeep and Ramalingam, Srikumar and Veit, Andreas and Glasner, Daniel and Chakrabarti, Ayan and Kumar, Sanjiv},
  booktitle={Proceedings of the IEEE/CVF Conference on Computer Vision and Pattern Recognition},
  pages={9307--9315},
  year={2024}
}

@inproceedings{fu2023dreamsim,
title={DreamSim: Learning New Dimensions of Human Visual Similarity using Synthetic Data},
author= {Fu, Stephanie and Tamir, Netanel and Sundaram, Shobhita and Chai, Lucy and Zhang, Richard and Dekel, Tali and Isola, Phillip},
booktitle={Advances in Neural Information Processing Systems},
pages={50742--50768},
volume={36},
year={2023}
}

@inproceedings{ke2021musiqmultiscaleimagequality,
  title={Musiq: Multi-scale image quality transformer},
  author={Ke, Junjie and Wang, Qifei and Wang, Yilin and Milanfar, Peyman and Yang, Feng},
  booktitle={Proceedings of the IEEE/CVF international conference on computer vision},
  pages={5148--5157},
  year={2021}
}

@ARTICLE{6353522,
  author={Mittal, Anish and Soundararajan, Rajiv and Bovik, Alan C.},
  journal={IEEE Signal Processing Letters}, 
  title={Making a “Completely Blind” Image Quality Analyzer}, 
  year={2013},
  volume={20},
  number={3},
  pages={209-212},
  doi={10.1109/LSP.2012.2227726}}

@article{wang2024exploiting,
  author = {Wang, Jianyi and Yue, Zongsheng and Zhou, Shangchen and Chan, Kelvin C.K. and Loy, Chen Change},
  title = {Exploiting Diffusion Prior for Real-World Image Super-Resolution},
  article = {International Journal of Computer Vision},
  year = {2024}
}

@inproceedings{liang2021swinir,
  title={Swinir: Image restoration using swin transformer},
  author={Liang, Jingyun and Cao, Jiezhang and Sun, Guolei and Zhang, Kai and Van Gool, Luc and Timofte, Radu},
  booktitle={Proceedings of the IEEE/CVF international conference on computer vision},
  pages={1833--1844},
  year={2021}
}

@article{roessle2023ganerf,
        title={GANeRF: Leveraging Discriminators to Optimize Neural Radiance Fields}, 
        author={Roessle, Barbara and M{\"u}ller, Norman and Porzi, Lorenzo and Bul{\`o}, Samuel Rota and Kontschieder, Peter and Nie{\ss}ner, Matthias},
        year = {2023},
        issue_date = {December 2023},
        publisher = {Association for Computing Machinery},
        address = {New York, NY, USA},
        volume = {42},
        number = {6},
        issn = {0730-0301},
        url = {https://doi.org/10.1145/3618402},
        doi = {10.1145/3618402},
        journal = {ACM Trans. Graph.},
        month = {nov},
        articleno = {207},
        numpages = {14},
}

@article{zwicker2002ewa,
  title={EWA splatting},
  author={Zwicker, Matthias and Pfister, Hanspeter and Van Baar, Jeroen and Gross, Markus},
  journal={IEEE Transactions on Visualization and Computer Graphics},
  volume={8},
  number={3},
  pages={223--238},
  year={2002},
  publisher={IEEE}
}
}
\clearpage
\appendix
\section{Appendix}

\subsection{Results for 8× Super-Resolution}
\label{sec:app:8x_sr}

Table~\ref{tab:tandt_8x} reports quantitative results at $8\times$ super-resolution (SR) for Tanks \& Temples~\cite{Knapitsch2017tandt}. We downsample the original $1920\times1080$ images by $16\times$ and upsample by $8\times$ to half the native resolution.
Our method achieves the best SSIM, PSNR, LPIPS, FID, and DreamSim scores. 3DGS (LR) attains higher MUSIQ and NIQE scores, consistent with the behavior discussed in Section~\ref{sec:results}. Our CMMD score ranks fourth among the compared methods. Prior work~\cite{jayasumana2024rethinkingfidbetterevaluation} shows that CMMD is sensitive to high-frequency distortions introduced by noise in the embedding space. We hypothesize that sharpening effects like mild aliasing or overly crisp edges at high upsampling factors may be interpreted as distortions by CMMD, even when they improve perceptual quality and are reflected positively by other metrics. As no single metric fully characterizes visual fidelity, we report a broad suite of metrics for a more complete evaluation.

Table~\ref{tab:deepblending_mipnerf_8x} presents the $8\times$ SR results for Deep Blending~\cite{DeepBlending2018} and Mip-NeRF 360~\cite{barron2022mipnerf360}. For Deep Blending, we downsample the original images by $8\times$ ($\sim$125$\times$125) and upsample by $8\times$ back to the native resolution ($\sim$1K$\times$1K). For Mip-NeRF 360, we downsample the original images by $16\times$ ($\sim$250$\times$188) and upsample by $8\times$ to half the native resolution ($\sim$2K$\times$1.5K). SplatSuRe achieves the best results across all metrics on Deep Blending and nearly all metrics on Mip-NeRF 360, slightly trailing only Mip-Splatting~\cite{Yu2024MipSplatting} on SSIM while outperforming SRGS~\cite{feng2024srgssuperresolution3dgaussian}. In these settings, the LR images contain limited high-frequency information due to their lower resolutions, requiring SR to hallucinate substantially more detail. This contrasts with the $4\times$ setting in Section~\ref{sec:results}, where the higher-resolution inputs in Mip-NeRF 360 reduce the need for SR and favor anti-aliasing approaches such as Mip-Splatting. These results emphasize that our selective SR method yields higher image quality than uniform application, even in settings with highly generative, view-inconsistent SR.

\subsection{Unified Training Pipeline}
\label{appendix:unified_pipeline}

Table~\ref{tab:unified_pipeline} evaluates a unified training pipeline that merges the LR initialization and SR refinement stages to avoid the training time overhead introduced by the two-stage pipeline in Section~\ref{sec:method}. The model is trained with LR images for the first 5K iterations to obtain stable geometry, after which the Gaussian fidelity scores and SR weight maps are computed and training continues with SR supervision for the remaining 25K iterations. The total budget of 30K iterations matches that used by baseline methods using a single-stage pipeline. This unified approach achieves performance comparable to the original two-stage formulation reported in Tables~\ref{tab:tandt} and \ref{tab:deepblending_mipnerf}, indicating that our SplatSuRe objective remains effective as a single continuous training schedule.

\subsection{Additional Ratio Threshold Analysis}
\label{appendix:per_scene_analysis}

Different scenes exhibit distinct behaviors as the ratio threshold and corresponding amount of super-resolution (SR) information increase. Most scenes benefit from a moderate amount of SR but experience a sharp drop in image quality when it is excessively applied, while others plateau or continue improving with diminishing returns. We visualize these trends by plotting PSNR and LPIPS across ratio thresholds for three representative scenes in each category. 

\begin{figure*}[!b]
    \centering
    \begin{subfigure}{0.32\linewidth}
        \centering
        \caption{\textbf{\emph{ballroom}} -- Tanks \& Temples~\cite{Knapitsch2017tandt}}
        \includegraphics[width=\textwidth]{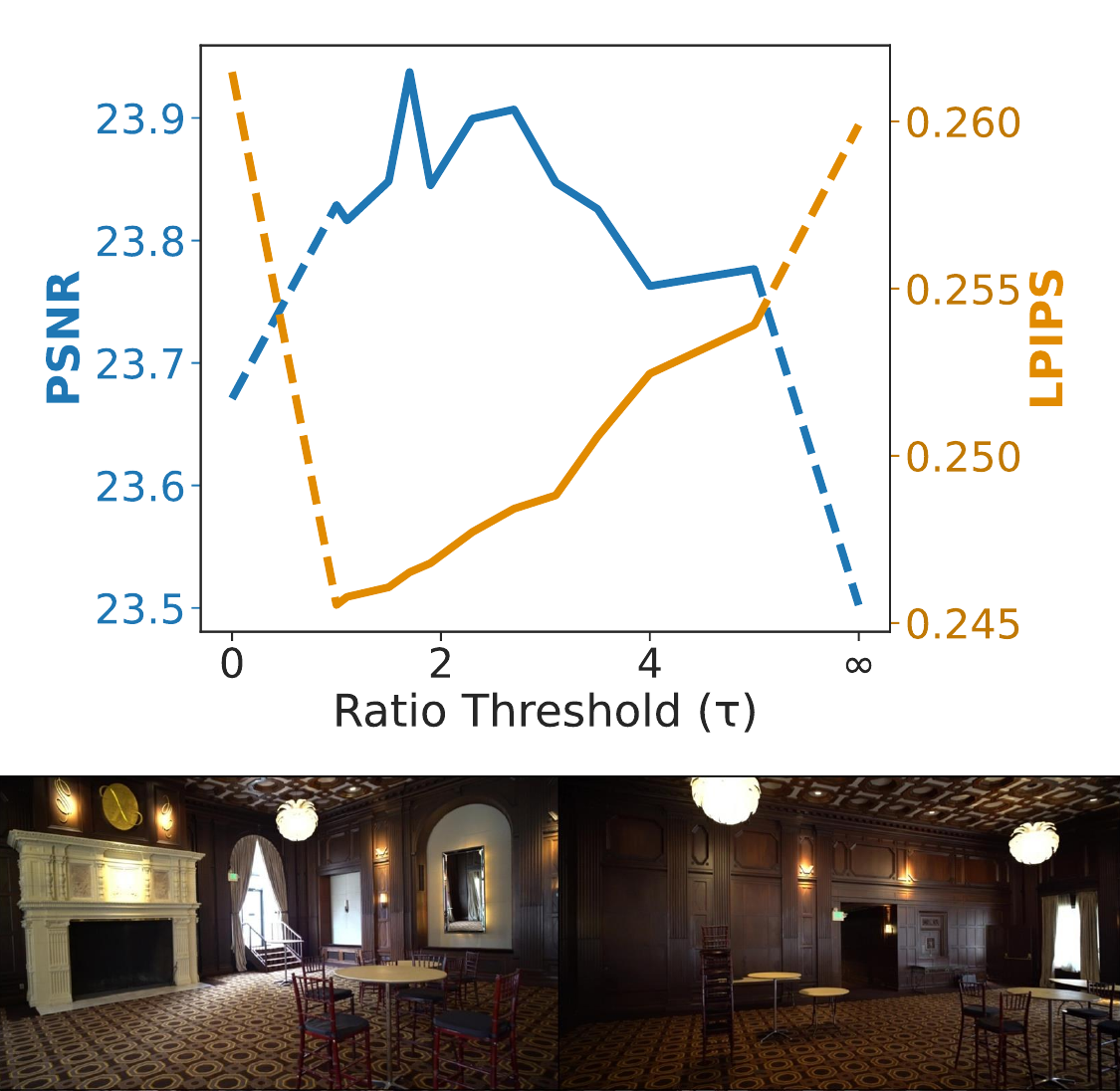}
    \end{subfigure}
    \begin{subfigure}{0.32\linewidth}
        \centering
        \caption{\textbf{\emph{kitchen}} -- Mip-NeRF 360~\cite{barron2022mipnerf360}}
        \includegraphics[width=\textwidth]{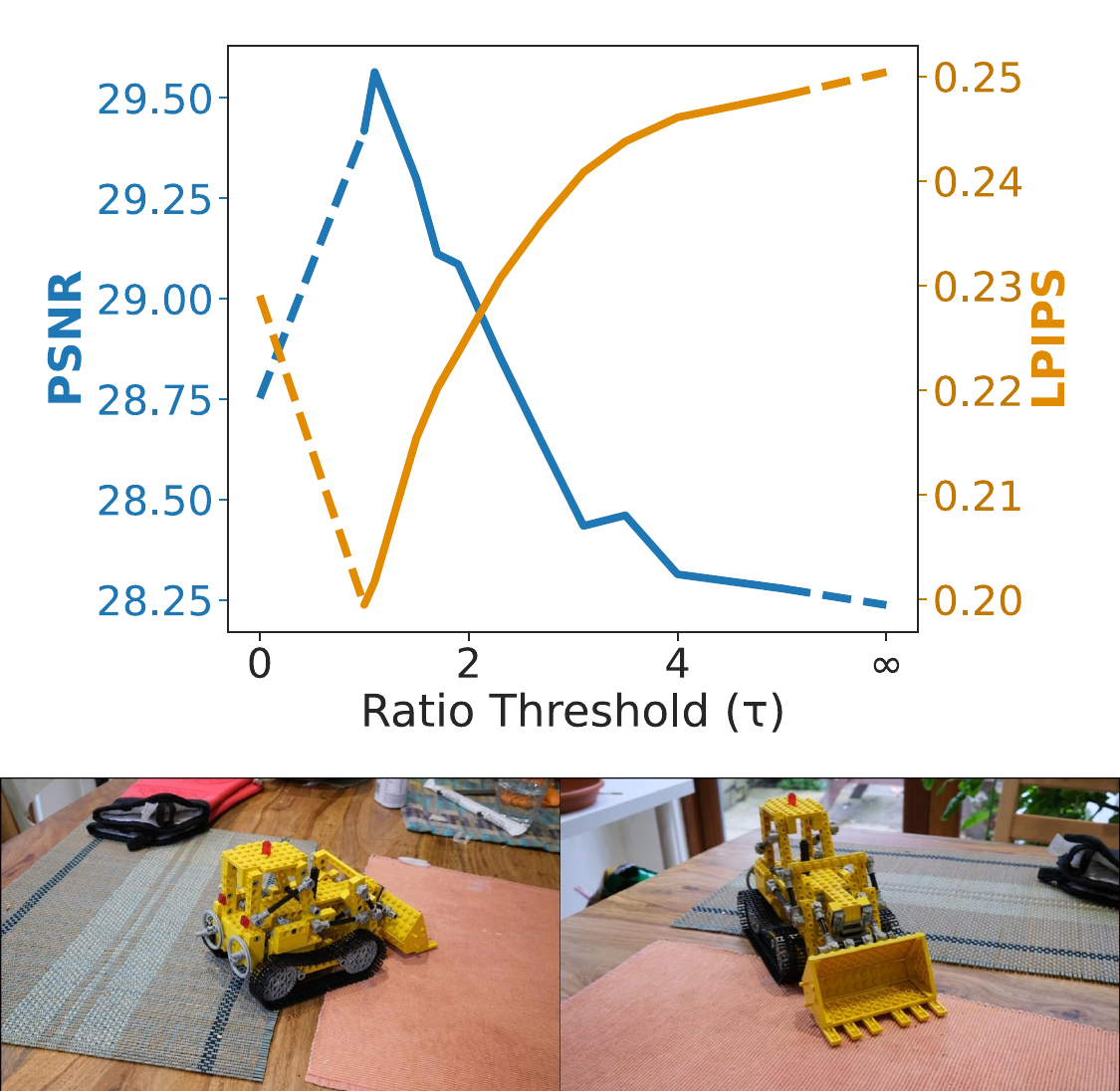}
    \end{subfigure}
    \begin{subfigure}{0.32\linewidth}
        \centering
        \caption{\textbf{\emph{francis}} -- Tanks \& Temples~\cite{Knapitsch2017tandt}}
        \includegraphics[width=\textwidth]{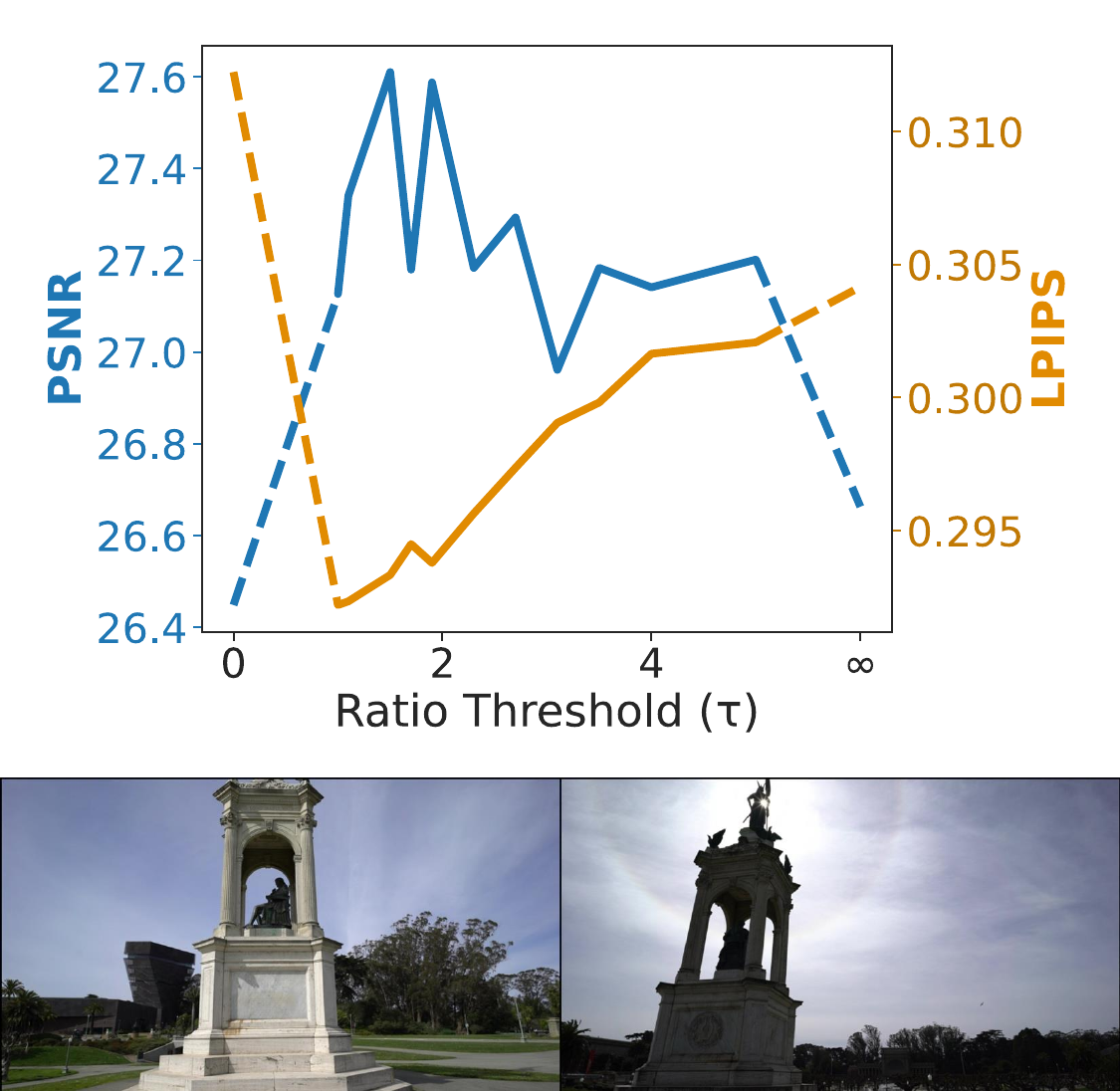}
    \end{subfigure}
    \caption{\textbf{Representative scenes that benefit from an optimal amount of super-resolution.} Top: Image quality vs. ratio threshold plots. Bottom: ground truth images illustrating scene structure for (a) \emph{ballroom} from Tanks \& Temples~\cite{Knapitsch2017tandt}, (b) \emph{kitchen} from Mip-NeRF 360~\cite{barron2022mipnerf360} and (c) \emph{francis} from Tanks \& Temples. Applying SR to the most poorly sampled regions yields large gains in image quality, whereas excessive SR introduces multi-view inconsistencies that sharply degrade quality. Most scenes exhibit this behavior, supporting our hypothesis that selectively applying SR is more beneficial than applying it uniformly.}
    \label{fig:optimal_sr_scenes}
\end{figure*}

Figure~\ref{fig:optimal_sr_scenes} illustrates scenes that benefit from an optimal amount of SR. Image quality initially improves with moderate SR but decreases sharply when it is excessively applied. Our method identifies the most poorly sampled regions and selectively applies SR to them, yielding a substantial initial quality boost. However, applying excessive SR introduces multi-view inconsistencies that rapidly degrade image quality. This behavior appears consistently across most scenes, supporting our hypothesis that selectively applying SR is more beneficial than applying it uniformly.

\begin{figure*}
    \centering
    \begin{subfigure}{0.32\linewidth}
        \centering
        \caption{\textbf{\emph{bicycle}} -- Mip-NeRF 360~\cite{barron2022mipnerf360}}
        \includegraphics[width=\textwidth]{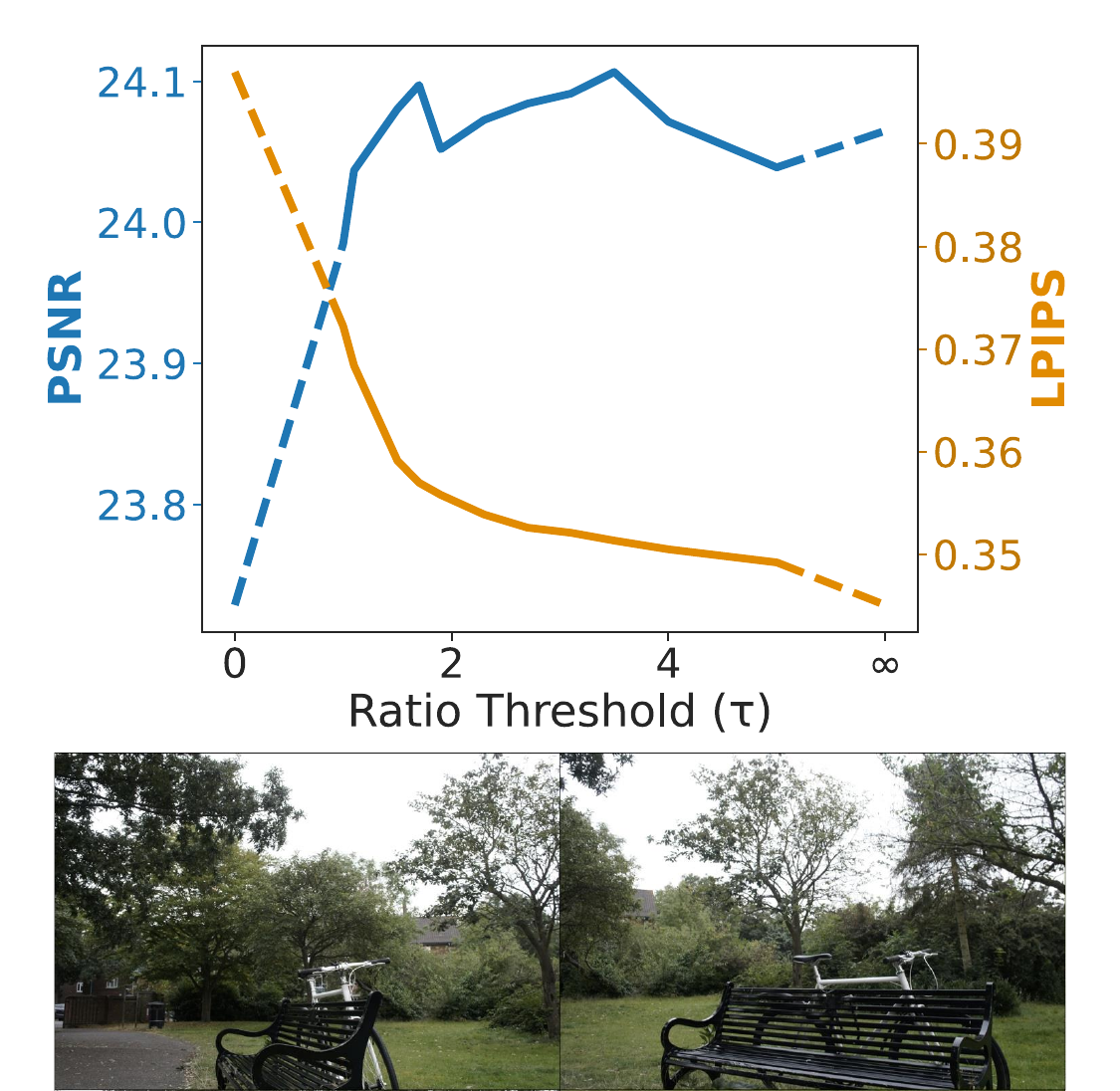}
    \end{subfigure}
    \begin{subfigure}{0.32\linewidth}
        \centering
        \caption{\textbf{\emph{garden}} -- Mip-NeRF 360~\cite{barron2022mipnerf360}}
        \includegraphics[width=\textwidth]{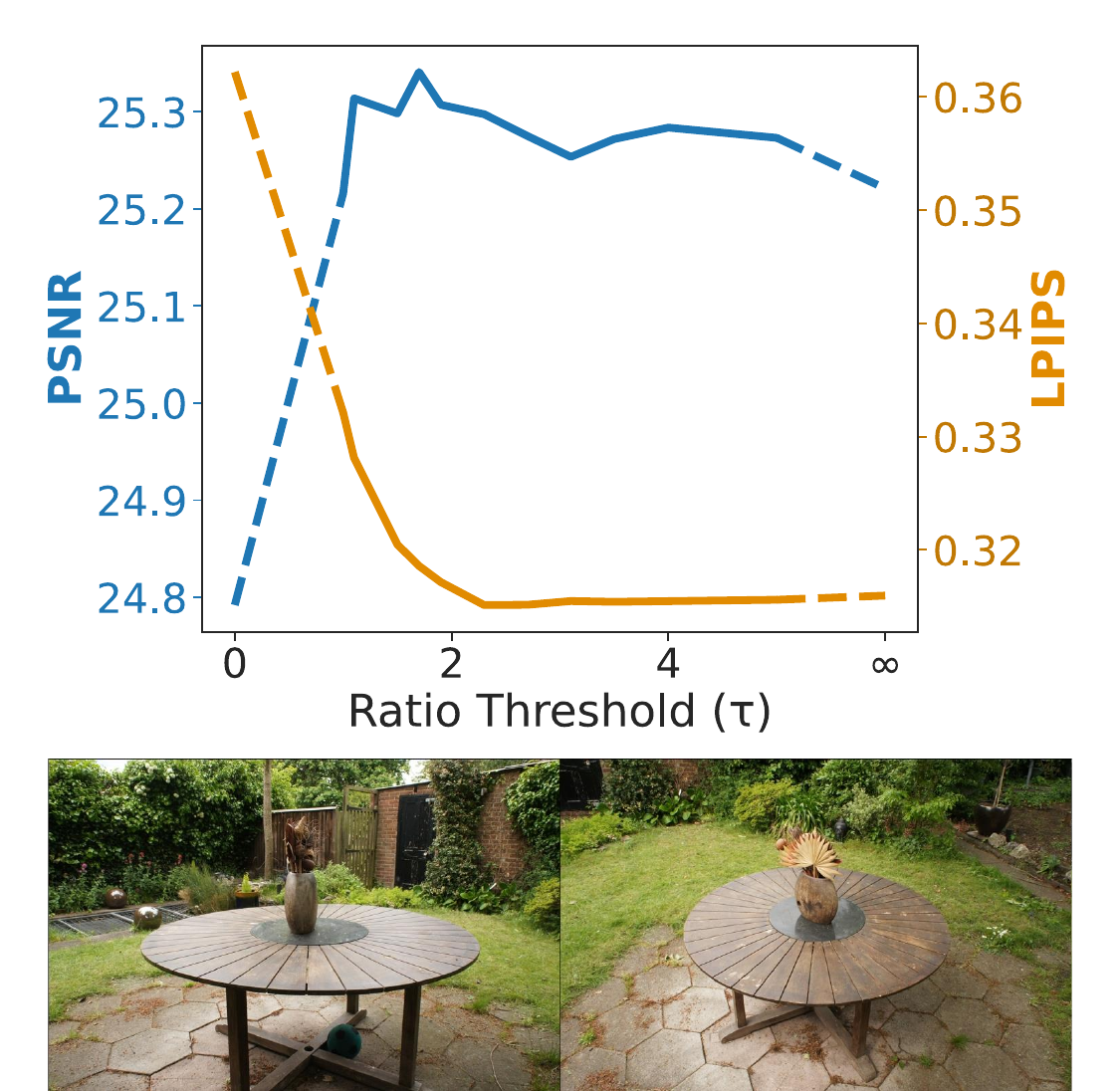}
    \end{subfigure}
    \begin{subfigure}{0.32\linewidth}
        \centering
        \caption{\textbf{\emph{stump}} -- Mip-NeRF 360~\cite{barron2022mipnerf360}}
        \includegraphics[width=\textwidth]{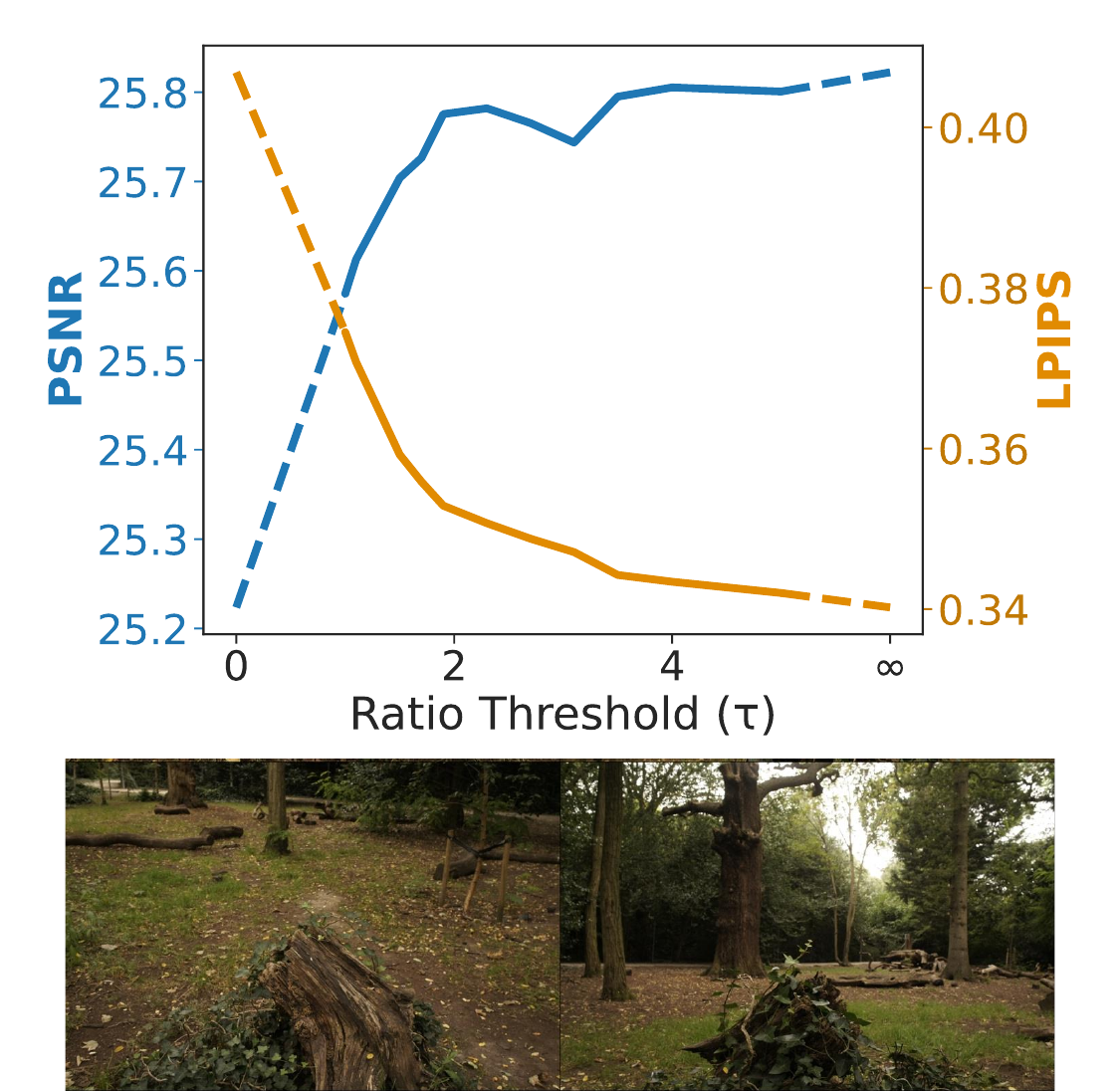}
    \end{subfigure}
    \caption{\textbf{Representative scenes that plateau in image quality or continue to benefit from increased amounts of super-resolution.} Top: Image quality vs. ratio threshold plots. Bottom: ground truth images illustrating scene structure for (a) \emph{bicycle}, (b) \emph{garden}, and (c) \emph{stump} from Mip-NeRF 360~\cite{barron2022mipnerf360}. Applying SR to the most poorly sampled regions yields large gains in image quality, while further increasing SR yields diminishing returns or no improvement. In particular, this occurs in scenes where the input images already contain substantial high-frequency detail and SR produces simpler sharpening or edge-enhancement effects rather than hallucinating new structure, making uniform application less harmful and sometimes marginally beneficial.}
    \label{fig:plateau_sr_scenes}
\end{figure*}

Figure~\ref{fig:plateau_sr_scenes} presents scenes that plateau in image quality or continue improving slightly as the amount of SR is increased. Our selective SR method produces sharp early quality gains at lower ratio thresholds, after which applying additional SR yields diminishing returns or no improvement. In particular, this occurs in scenes where the input images already contain substantial high-frequency detail and SR produces simpler sharpening or edge enhancement effects rather than hallucinating new structure, making uniform application less harmful and sometimes marginally beneficial. This behavior is especially common in outdoor scenes in Mip-NeRF 360~\cite{barron2022mipnerf360}, where the downsampled $\sim500\times375$ images remain relatively high-resolution and therefore do not exhibit the multi-view inconsistencies that typically arise when excessive SR is applied.

\subsection{Additional Visualizations}
\label{appendix:additional_results}

\begin{figure*}
    \centering
    \includegraphics[width=\linewidth]{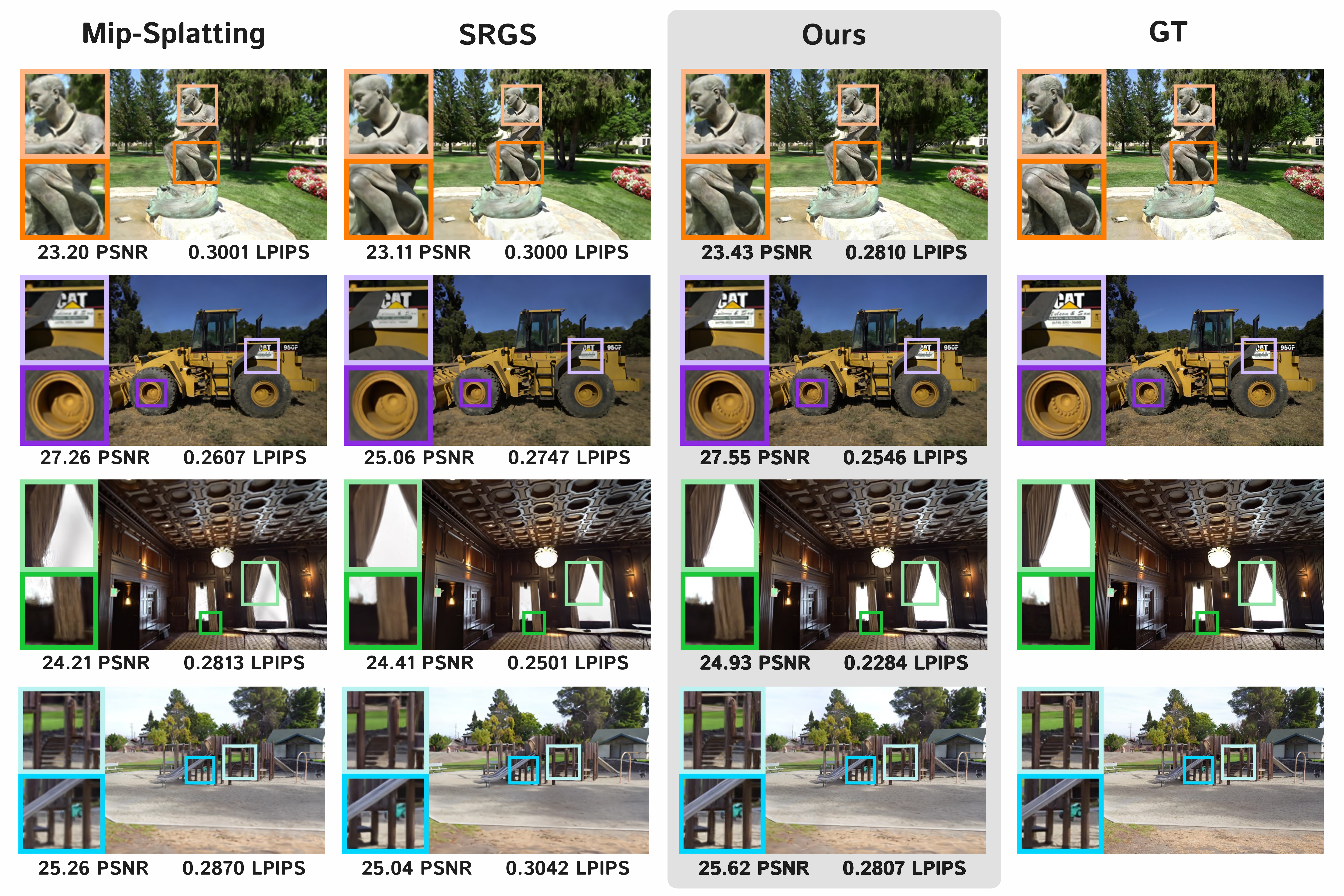}
    \caption{\textbf{Additional qualitative results on Tanks \& Temples~\cite{Knapitsch2017tandt}.} Experiments are performed at $4\times$ super-resolution with ratio threshold $\tau{=}1.1$. Compared to Mip-Splatting~\cite{Yu2024MipSplatting} and SRGS~\cite{feng2024srgssuperresolution3dgaussian}, our method produces sharper, more faithful reconstructions that better align with ground truth while maintaining cross-view consistency. It preserves high-frequency patterns (\ctext[RGB]{255, 177, 115}{orange boxes} on statue), fine details in text (\ctext[RGB]{210, 162, 247}{purple box} on vehicle), and reduces artifacts while preserving sharpness (\ctext[RGB]{72, 203, 97}{green boxes} on curtains, \ctext[RGB]{183, 240, 241}{blue boxes} in playground).}
    \label{fig:supplementary_qualitative_1}
\end{figure*}

\begin{figure*}
    \centering
    \includegraphics[width=\linewidth]{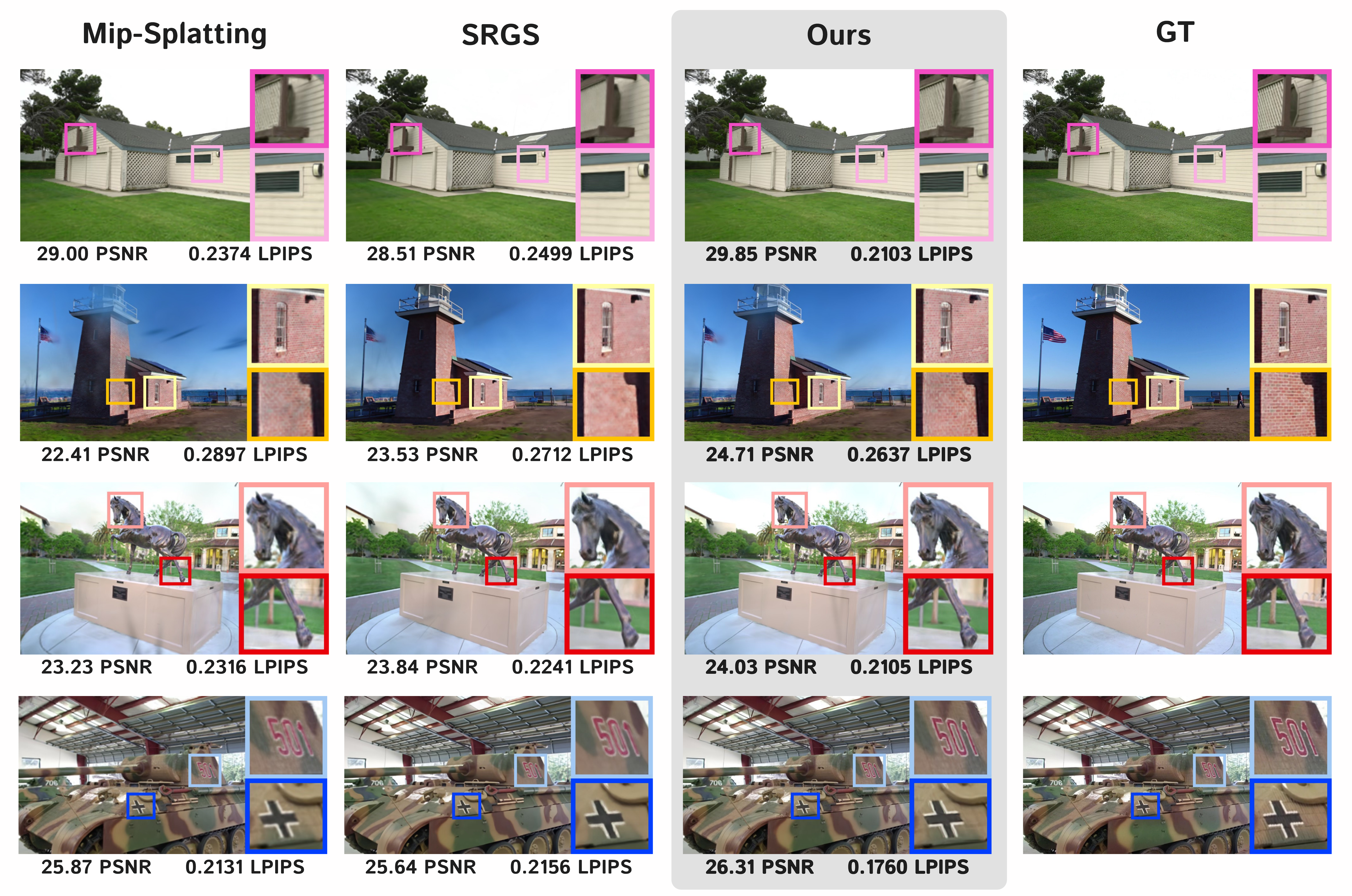}
    \caption{\textbf{Additional qualitative results on Tanks \& Temples~\cite{Knapitsch2017tandt}.} Experiments are performed at $4\times$ super-resolution with ratio threshold $\tau{=}1.1$. Compared to Mip-Splatting~\cite{Yu2024MipSplatting} and SRGS~\cite{feng2024srgssuperresolution3dgaussian}, our method produces sharper, more faithful reconstructions that better align with the ground truth while maintaining cross-view consistency. It preserves high-frequency geometry (\ctext[RGB]{253, 172, 231}{pink boxes} on barn) and retains sharp texture (\ctext[RGB]{255, 196, 0}{yellow boxes} on lighthouse, \ctext[RGB]{255, 89, 89}{red boxes} on horse statue, \ctext[RGB]{183, 240, 241}{blue boxes} on tank).}
    \label{fig:supplementary_qualitative_2}
\end{figure*}

Figures~\ref{fig:supplementary_qualitative_1} and \ref{fig:supplementary_qualitative_2} present additional qualitative results on Tanks \& Temples~\cite{Knapitsch2017tandt}. Consistent with the examples in Figures~\ref{fig:teaser} and \ref{fig:main_results_fig} discussed in Section~\ref{sec:results}, SplatSuRe produces sharper reconstructions than competing methods while reducing artifacts and preserving smoothness in uniformly textured regions.

\subsection{Per-Scene Metrics}
\label{sec:app:scene_metrics}

Tables~\ref{tab:ssim}, \ref{tab:psnr}, \ref{tab:lpips}, \ref{tab:fid}, \ref{tab:cmmd}, \ref{tab:dreamsim}, \ref{tab:musiq}, and \ref{tab:niqe} report per-scene SSIM, PSNR, LPIPS, FID, CMMD, DreamSim, MUSIQ, and NIQE results on Tanks \& Temples~\cite{Knapitsch2017tandt}, Deep Blending~\cite{DeepBlending2018}, and Mip-NeRF 360~\cite{barron2022mipnerf360} for all methods evaluated in Section~\ref{sec:experiments}. Across individual scenes, we observe the same trends as in the averaged results presented by Tables~\ref{tab:tandt} and \ref{tab:deepblending_mipnerf} in Section~\ref{sec:results}: SplatSuRe achieves the strongest performance on Tanks \& Temples and Deep Blending and outperforms SRGS on Mip-NeRF 360.


\begin{table*}[t]
\centering
\caption{\textbf{Quantitative results on Tanks \& Temples~\cite{Knapitsch2017tandt} at 8$\times$ super-resolution.} Experiments are performed using ratio threshold $\tau{=}1.1$. The \metrictablebest{best}, \metrictablesecond{second best} and \metrictablethird{third best} entries are highlighted. Our SplatSuRe method achieves the strongest results on most metrics.}
\resizebox{\textwidth}{!}{
\begin{tabular}{l|cccccccccccc}
\toprule
\multirow{2}{*}{\textbf{Method}} & \multicolumn{8}{c}{\textbf{Tanks \& Temples~\cite{Knapitsch2017tandt}}} \\
& SSIM~$\uparrow$ & PSNR~$\uparrow$ & LPIPS~$\downarrow$ & FID~$\downarrow$ & CMMD~$\downarrow$ & DreamSim~$\downarrow$ & MUSIQ~$\uparrow$ & NIQE~$\downarrow$\\
\midrule
3DGS (LR)~\cite{kerbl3Dgaussians} & 
0.537 & 16.40 & 0.473 & 172.44 & 3.795 & 0.2427 & \metrictablebest{46.870} & \metrictablebest{3.976} \\
3DGS~\cite{kerbl3Dgaussians} + StableSR~\cite{wang2024exploiting} & \metrictablethird{0.664} & 21.49 & \metrictablethird{0.406} & \metrictablethird{99.80} & \metrictablebest{1.878} & \metrictablethird{0.1100} & \metrictablethird{38.929} & 6.293 \\
Mip-Splatting~\cite{Yu2024MipSplatting} & 0.661 & \metrictablethird{21.53} & 0.428 & 109.01 & \metrictablethird{1.941} & 0.1183 & 32.417 & 6.645 \\
SRGS~\cite{feng2024srgssuperresolution3dgaussian} + StableSR~\cite{wang2024exploiting} & \metrictablesecond{0.674} & \metrictablesecond{21.97} & \metrictablesecond{0.399} & \metrictablesecond{92.91} & \metrictablesecond{1.891} & \metrictablesecond{0.0981} & 38.322 & \metrictablethird{6.138} \\
Ours + StableSR~\cite{wang2024exploiting} & \metrictablebest{0.692} & \metrictablebest{22.48} & \metrictablebest{0.378} & \metrictablebest{77.17} & 2.007 & \metrictablebest{0.0774} & \metrictablesecond{44.428} & \metrictablesecond{5.341} \\
\bottomrule
\end{tabular}
}
\label{tab:tandt_8x}
\end{table*}

\begin{table*}[t]
\centering
\caption{\textbf{Quantitative results on Deep Blending~\cite{DeepBlending2018} and Mip-NeRF 360~\cite{barron2022mipnerf360} at 8$\times$ super-resolution.} Experiments are performed using ratio threshold $\tau{=}1.1$. Our SplatSuRe method achieves the strongest results on almost all metrics.}
\resizebox{\textwidth}{!}{
\begin{tabular}{l|ccccc|ccccc}
\toprule
\multirow{2}{*}{\textbf{Method}} & \multicolumn{5}{c|}{\textbf{Deep Blending~\cite{DeepBlending2018}}} & \multicolumn{5}{c}{\textbf{Mip-NeRF 360~\cite{barron2022mipnerf360}}}\\
&SSIM~$\uparrow$ & PSNR~$\uparrow$ & LPIPS~$\downarrow$ & CMMD~$\downarrow$&DreamSim~$\downarrow$&SSIM~$\uparrow$ & PSNR~$\uparrow$ & LPIPS~$\downarrow$ & CMMD~$\downarrow$&DreamSim~$\downarrow$\\
\midrule
3DGS (LR)~\cite{kerbl3Dgaussians} & 0.783 & 24.50 & 0.404 & 2.051 & 0.1280 & 0.509 & 18.17 & 0.491 & 2.395 & 0.1337 \\
3DGS~\cite{kerbl3Dgaussians} + StableSR~\cite{wang2024exploiting} & 0.821 & 26.68 & \metrictablethird{0.383} & 1.159 & 0.0705 & 0.630 & 24.23 & 0.445 & 1.096 & 0.0546 \\
Mip-Splatting~\cite{Yu2024MipSplatting} & \metrictablesecond{0.831} & \metrictablethird{27.24} & 0.391 & \metrictablethird{1.127} & \metrictablethird{0.0689} & \metrictablebest{0.656} & \metrictablethird{24.80} & \metrictablesecond{0.420} & \metrictablesecond{0.610} & \metrictablesecond{0.0243} \\
SRGS~\cite{feng2024srgssuperresolution3dgaussian} + StableSR~\cite{wang2024exploiting} & \metrictablethird{0.830} & \metrictablesecond{27.35} & \metrictablesecond{0.377} & \metrictablesecond{1.079} & \metrictablesecond{0.0622} & \metrictablethird{0.648} & \metrictablesecond{24.88} & \metrictablethird{0.422} & \metrictablethird{0.665} & \metrictablethird{0.0320} \\
Ours + StableSR~\cite{wang2024exploiting} & \metrictablebest{0.843} & \metrictablebest{28.03} & \metrictablebest{0.357} & \metrictablebest{0.879} & \metrictablebest{0.0489} & \metrictablesecond{0.655} & \metrictablebest{25.08} & \metrictablebest{0.406} & \metrictablebest{0.546} & \metrictablebest{0.0231} \\
\bottomrule
\end{tabular}
}
\label{tab:deepblending_mipnerf_8x}
\end{table*}

\begin{table*}[t]
\centering
\caption{\textbf{Quantitative comparison of our unified and two-stage pipelines at $\mathbf{4\times}$ SR across Tanks \& Temples~\cite{Knapitsch2017tandt}, Deep Blending~\cite{DeepBlending2018}, and Mip-NeRF 360~\cite{barron2022mipnerf360}.} Experiments are performed using ratio threshold $\tau=1.1$. The best entry is \textbf{bolded}. The unified pipeline achieves similar performance to the two-stage approach while requiring less training time.}
\resizebox{\linewidth}{!}{
\begin{tabular}{ll|cccccccc}
\toprule
\textbf{Dataset} & \textbf{Method} & SSIM~$\uparrow$ & PSNR~$\uparrow$ & LPIPS~$\downarrow$ & FID~$\downarrow$ & CMMD~$\downarrow$ & DreamSim~$\downarrow$ & MUSIQ~$\uparrow$ & NIQE~$\downarrow$ \\
\midrule
\multirow{2}{*}{Tanks \& Temples~\cite{Knapitsch2017tandt}}
& Two-Stage & \textbf{0.784} & \textbf{23.81} & 0.272 & 37.72 & 1.040 & 0.0413 & 58.332 & 3.928 \\
& Unified & 0.781 & 23.70 & \textbf{0.271} & \textbf{36.02} & \textbf{1.006} & \textbf{0.0400} & \textbf{59.701} & \textbf{3.705} \\
\midrule
\multirow{2}{*}{Deep Blending~\cite{DeepBlending2018}}
& Two-Stage & \textbf{0.872} & \textbf{29.01} & 0.306 & 44.14 & 0.496 & 0.0330 & \textbf{53.019} & 5.411\\
& Unified & 0.871 & 28.96 & \textbf{0.304} & \textbf{41.70} & \textbf{0.483} & \textbf{0.0326} & 52.983 & \textbf{5.286}\\
\midrule
\multirow{2}{*}{Mip-NeRF 360~\cite{barron2022mipnerf360}}
& Two-Stage & 0.740 & 26.34 & 0.323 & \textbf{27.56} & 0.339 & 0.0179  & 54.366 & 3.934\\
& Unified & \textbf{0.758} & \textbf{26.69} & \textbf{0.304} & 26.18 & \textbf{0.271} & \textbf{0.0164} & \textbf{56.773} & \textbf{3.715} \\
\bottomrule
\end{tabular}
}
\label{tab:unified_pipeline}
\end{table*}

\clearpage

\clearpage

\clearpage

\begin{table*}[t]
\centering
\caption{\textbf{SSIM~$\uparrow$ on each scene in Tanks \& Temples~\cite{Knapitsch2017tandt}, Deep Blending~\cite{DeepBlending2018}, and Mip-NeRF 360~\cite{barron2022mipnerf360}.} Experiments are performed at \textbf{$4\times$} super-resolution using ratio threshold $\tau{=}1.1$. The \metrictablebest{best}, \metrictablesecond{second best} and \metrictablethird{third best} entries are highlighted.}
\resizebox{\textwidth}{!}{
\begin{tabular}{l|cccccccccccccccccccc}
\toprule
\multirow{2}{*}{\textbf{Method}} & \multicolumn{19}{c}{\textbf{Tanks \& Temples~\cite{Knapitsch2017tandt}}} \\
& \textit{auditorium} & \textit{ignatius} & \textit{palace} & \textit{ballroom} & \textit{panther} & \textit{barn} & \textit{lighthouse} & \textit{playground} & \textit{courtroom} & \textit{m60} & \textit{temple} & \textit{caterpillar} & \textit{family} & \textit{train} & \textit{francis} & \textit{truck} & \textit{church} & \textit{horse} & \textit{museum} \\
\midrule
3DGS (LR)~\cite{kerbl3Dgaussians} & 0.807 & 0.527 & 0.666 & 0.578 & 0.752 & 0.729 & 0.742 & 0.628 & 0.647 & 0.735 & 0.715 & 0.602 & 0.566 & 0.648 & 0.779 & 0.649 & 0.682 & 0.675 & 0.579 \\
3DGS~\cite{kerbl3Dgaussians} + StableSR~\cite{wang2024exploiting} & \metrictablethird{0.849} & 0.646 & \metrictablethird{0.709} & 0.702 & 0.819 & 0.771 & 0.777 & 0.720 & 0.725 & 0.810 & 0.761 & 0.686 & 0.753 & 0.705 & 0.845 & 0.759 & 0.744 & 0.811 & 0.683 \\
Mip-Splatting~\cite{Yu2024MipSplatting} & 0.838 & \metrictablesecond{0.681} & 0.687 & \metrictablethird{0.729} & \metrictablesecond{0.847} & \metrictablethird{0.790} & \metrictablethird{0.789} & \metrictablesecond{0.746} & \metrictablethird{0.730} & \metrictablesecond{0.836} & \metrictablethird{0.762} & \metrictablesecond{0.704} & \metrictablesecond{0.788} & \metrictablesecond{0.733} & \metrictablethird{0.855} & \metrictablesecond{0.786} & \metrictablethird{0.753} & \metrictablethird{0.827} & \metrictablethird{0.689} \\
SRGS~\cite{feng2024srgssuperresolution3dgaussian} + StableSR~\cite{wang2024exploiting} & \metrictablebest{0.861} & \metrictablethird{0.671} & \metrictablesecond{0.713} & \metrictablesecond{0.734} & \metrictablethird{0.846} & \metrictablesecond{0.794} & \metrictablesecond{0.789} & \metrictablethird{0.742} & \metrictablesecond{0.746} & \metrictablethird{0.835} & \metrictablesecond{0.773} & \metrictablethird{0.703} & \metrictablethird{0.780} & \metrictablethird{0.729} & \metrictablesecond{0.857} & \metrictablethird{0.784} & \metrictablesecond{0.761} & \metrictablesecond{0.829} & \metrictablesecond{0.704} \\
Ours + StableSR~\cite{wang2024exploiting} & \metrictablesecond{0.859} & \metrictablebest{0.694} & \metrictablebest{0.717} & \metrictablebest{0.746} & \metrictablebest{0.859} & \metrictablebest{0.808} & \metrictablebest{0.807} & \metrictablebest{0.764} & \metrictablebest{0.752} & \metrictablebest{0.846} & \metrictablebest{0.786} & \metrictablebest{0.714} & \metrictablebest{0.795} & \metrictablebest{0.743} & \metrictablebest{0.866} & \metrictablebest{0.798} & \metrictablebest{0.781} & \metrictablebest{0.835} & \metrictablebest{0.716} \\
\bottomrule
\end{tabular}
}
\resizebox{0.6\textwidth}{!}{
\begin{tabular}{l|cc|cccccccccc}
\toprule
\multirow{2}{*}{\textbf{Method}} & \multicolumn{2}{c|}{\textbf{Deep Blending~\cite{DeepBlending2018}}} & \multicolumn{9}{c}{\textbf{Mip-NeRF 360~\cite{barron2022mipnerf360}}} \\
& \textit{drjohnson} & \textit{playroom} & \textit{bicycle} & \textit{bonsai} & \textit{counter} & \textit{flowers} & \textit{garden} & \textit{kitchen} & \textit{room} & \textit{stump} & \textit{treehill} \\
\midrule
3DGS (LR)~\cite{kerbl3Dgaussians} & 0.828 & 0.843 & 0.527 & 0.776 & 0.772 & 0.429 & 0.559 & 0.719 & 0.836 & 0.581 & 0.516 \\
3DGS~\cite{kerbl3Dgaussians} + StableSR~\cite{wang2024exploiting} & 0.837 & 0.854 & 0.604 & 0.848 & 0.833 & 0.497 & 0.654 & 0.751 & 0.858 & 0.669 & 0.571 \\
Mip-Splatting~\cite{Yu2024MipSplatting} & \metrictablesecond{0.855} & \metrictablesecond{0.874} & \metrictablebest{0.676} & \metrictablebest{0.917} & \metrictablebest{0.872} & \metrictablebest{0.544} & \metrictablebest{0.719} & \metrictablebest{0.870} & \metrictablebest{0.897} & \metrictablebest{0.740} & \metrictablebest{0.596} \\
SRGS~\cite{feng2024srgssuperresolution3dgaussian} + StableSR~\cite{wang2024exploiting} & \metrictablethird{0.851} & \metrictablethird{0.870} & \metrictablesecond{0.653} & \metrictablethird{0.896} & \metrictablethird{0.863} & \metrictablesecond{0.533} & \metrictablesecond{0.692} & \metrictablethird{0.825} & \metrictablethird{0.888} & \metrictablethird{0.671} & \metrictablesecond{0.590} \\
Ours + StableSR~\cite{wang2024exploiting} & \metrictablebest{0.867} & \metrictablebest{0.877} & \metrictablethird{0.646} & \metrictablesecond{0.906} & \metrictablesecond{0.867} & \metrictablethird{0.503} & \metrictablethird{0.690} & \metrictablesecond{0.869} & \metrictablesecond{0.896} & \metrictablesecond{0.706} & \metrictablethird{0.582} \\
\bottomrule
\end{tabular}
}
\label{tab:ssim}
\end{table*}

\begin{table*}[t]
\centering
\caption{\textbf{PSNR~$\uparrow$ on each scene in Tanks \& Temples~\cite{Knapitsch2017tandt}, Deep Blending~\cite{DeepBlending2018}, and Mip-NeRF 360~\cite{barron2022mipnerf360}.}}
\resizebox{\textwidth}{!}{
\begin{tabular}{l|cccccccccccccccccccc}
\toprule
\multirow{2}{*}{\textbf{Method}} & \multicolumn{19}{c}{\textbf{Tanks \& Temples~\cite{Knapitsch2017tandt}}} \\
& \textit{auditorium} & \textit{ignatius} & \textit{palace} & \textit{ballroom} & \textit{panther} & \textit{barn} & \textit{lighthouse} & \textit{playground} & \textit{courtroom} & \textit{m60} & \textit{temple} & \textit{caterpillar} & \textit{family} & \textit{train} & \textit{francis} & \textit{truck} & \textit{church} & \textit{horse} & \textit{museum} \\
\midrule
3DGS (LR)~\cite{kerbl3Dgaussians} & 22.14 & 17.13 & 18.35 & 18.23 & 22.49 & 22.97 & 20.47 & 19.85 & 19.53 & 21.72 & 18.99 & 19.36 & 15.09 & 18.80 & 22.47 & 17.38 & 20.00 & 16.30 & 17.44 \\
3DGS~\cite{kerbl3Dgaussians} + StableSR~\cite{wang2024exploiting} & \metrictablethird{23.62} & 20.52 & \metrictablethird{19.68} & 22.20 & 25.23 & 25.18 & \metrictablethird{21.65} & 23.24 & 21.75 & 24.72 & 20.24 & 22.44 & 22.54 & 20.55 & 25.97 & 22.86 & \metrictablethird{22.02} & 22.72 & 19.76 \\
Mip-Splatting~\cite{Yu2024MipSplatting} & 23.46 & \metrictablethird{21.22} & 18.24 & \metrictablethird{23.15} & \metrictablesecond{27.11} & \metrictablesecond{26.85} & 21.29 & \metrictablesecond{24.13} & \metrictablethird{22.26} & \metrictablethird{26.16} & \metrictablethird{20.31} & \metrictablesecond{22.99} & \metrictablebest{24.09} & \metrictablesecond{21.33} & \metrictablesecond{26.62} & \metrictablesecond{23.94} & 21.99 & \metrictablethird{23.48} & \metrictablethird{20.28} \\
SRGS~\cite{feng2024srgssuperresolution3dgaussian} + StableSR~\cite{wang2024exploiting} & \metrictablesecond{24.23} & \metrictablesecond{21.26} & \metrictablesecond{19.69} & \metrictablesecond{23.44} & \metrictablethird{26.98} & \metrictablethird{26.83} & \metrictablesecond{21.74} & \metrictablethird{24.12} & \metrictablesecond{22.74} & \metrictablesecond{26.17} & \metrictablesecond{20.49} & \metrictablethird{22.98} & \metrictablethird{23.82} & \metrictablethird{21.28} & \metrictablethird{26.55} & \metrictablethird{23.87} & \metrictablesecond{22.54} & \metrictablesecond{23.72} & \metrictablesecond{20.63} \\
Ours + StableSR~\cite{wang2024exploiting} & \metrictablebest{24.66} & \metrictablebest{21.61} & \metrictablebest{20.24} & \metrictablebest{23.82} & \metrictablebest{27.38} & \metrictablebest{27.33} & \metrictablebest{22.82} & \metrictablebest{25.12} & \metrictablebest{22.99} & \metrictablebest{26.63} & \metrictablebest{21.39} & \metrictablebest{23.34} & \metrictablesecond{24.00} & \metrictablebest{21.79} & \metrictablebest{27.34} & \metrictablebest{24.11} & \metrictablebest{23.18} & \metrictablebest{23.94} & \metrictablebest{20.77} \\
\bottomrule
\end{tabular}
}
\resizebox{0.6\textwidth}{!}{
\begin{tabular}{l|cc|cccccccccc}
\toprule
\multirow{2}{*}{\textbf{Method}} & \multicolumn{2}{c|}{\textbf{Deep Blending~\cite{DeepBlending2018}}} & \multicolumn{9}{c}{\textbf{Mip-NeRF 360~\cite{barron2022mipnerf360}}} \\
& \textit{drjohnson} & \textit{playroom} & \textit{bicycle} & \textit{bonsai} & \textit{counter} & \textit{flowers} & \textit{garden} & \textit{kitchen} & \textit{room} & \textit{stump} & \textit{treehill} \\
\midrule
3DGS (LR)~\cite{kerbl3Dgaussians} & 26.29 & 27.15 & 18.50 & 22.75 & 23.27 & 17.60 & 18.64 & 19.55 & 25.93 & 20.34 & 19.44 \\
3DGS~\cite{kerbl3Dgaussians} + StableSR~\cite{wang2024exploiting} & 26.60 & 27.71 & 22.87 & 26.48 & 26.44 & 19.92 & 23.80 & 24.80 & 28.32 & 24.24 & 21.64 \\
Mip-Splatting~\cite{Yu2024MipSplatting} & \metrictablesecond{27.54} & \metrictablesecond{29.33} & \metrictablebest{24.28} & \metrictablebest{30.65} & \metrictablesecond{28.28} & \metrictablebest{21.05} & \metrictablebest{25.63} & \metrictablesecond{29.48} & \metrictablesecond{30.69} & \metrictablebest{26.20} & \metrictablebest{22.08} \\
SRGS~\cite{feng2024srgssuperresolution3dgaussian} + StableSR~\cite{wang2024exploiting} & \metrictablethird{27.38} & \metrictablethird{29.08} & \metrictablesecond{24.05} & \metrictablethird{29.81} & \metrictablethird{28.11} & \metrictablesecond{20.91} & \metrictablethird{25.20} & \metrictablethird{28.26} & \metrictablethird{30.42} & \metrictablethird{24.58} & \metrictablethird{21.97} \\
Ours + StableSR~\cite{wang2024exploiting} & \metrictablebest{28.46} & \metrictablebest{29.57} & \metrictablethird{24.04} & \metrictablesecond{30.54} & \metrictablebest{28.39} & \metrictablethird{20.55} & \metrictablesecond{25.31} & \metrictablebest{29.56} & \metrictablebest{31.05} & \metrictablesecond{25.61} & \metrictablesecond{22.02} \\
\bottomrule
\end{tabular}
}
\label{tab:psnr}
\end{table*}

\begin{table*}[t]
\centering
\caption{\textbf{LPIPS~$\downarrow$ on each scene in Tanks \& Temples~\cite{Knapitsch2017tandt}, Deep Blending~\cite{DeepBlending2018}, and Mip-NeRF 360~\cite{barron2022mipnerf360}.}}
\resizebox{\textwidth}{!}{
\begin{tabular}{l|cccccccccccccccccccc}
\toprule
\multirow{2}{*}{\textbf{Method}} & \multicolumn{19}{c}{\textbf{Tanks \& Temples~\cite{Knapitsch2017tandt}}} \\
& \textit{auditorium} & \textit{ignatius} & \textit{palace} & \textit{ballroom} & \textit{panther} & \textit{barn} & \textit{lighthouse} & \textit{playground} & \textit{courtroom} & \textit{m60} & \textit{temple} & \textit{caterpillar} & \textit{family} & \textit{train} & \textit{francis} & \textit{truck} & \textit{church} & \textit{horse} & \textit{museum} \\
\midrule
3DGS (LR)~\cite{kerbl3Dgaussians} & 0.301 & 0.411 & 0.380 & 0.337 & 0.309 & 0.317 & 0.303 & 0.371 & 0.376 & 0.302 & 0.335 & 0.385 & 0.408 & 0.346 & 0.352 & 0.345 & 0.343 & 0.322 & 0.400 \\
3DGS~\cite{kerbl3Dgaussians} + StableSR~\cite{wang2024exploiting} & \metrictablesecond{0.247} & 0.357 & \metrictablesecond{0.364} & \metrictablethird{0.283} & 0.275 & 0.295 & \metrictablethird{0.283} & 0.333 & \metrictablethird{0.294} & 0.266 & \metrictablethird{0.302} & 0.340 & 0.289 & 0.305 & 0.316 & 0.298 & \metrictablethird{0.306} & \metrictablethird{0.234} & \metrictablethird{0.317} \\
Mip-Splatting~\cite{Yu2024MipSplatting} & 0.275 & \metrictablethird{0.331} & 0.408 & 0.288 & \metrictablethird{0.260} & \metrictablethird{0.293} & 0.296 & \metrictablethird{0.316} & 0.323 & \metrictablethird{0.250} & 0.325 & \metrictablethird{0.325} & \metrictablethird{0.281} & \metrictablethird{0.298} & \metrictablethird{0.316} & \metrictablethird{0.284} & 0.313 & 0.237 & 0.345 \\
SRGS~\cite{feng2024srgssuperresolution3dgaussian} + StableSR~\cite{wang2024exploiting} & \metrictablebest{0.242} & \metrictablesecond{0.328} & \metrictablethird{0.367} & \metrictablesecond{0.266} & \metrictablesecond{0.251} & \metrictablesecond{0.277} & \metrictablesecond{0.276} & \metrictablesecond{0.311} & \metrictablesecond{0.292} & \metrictablesecond{0.242} & \metrictablesecond{0.299} & \metrictablesecond{0.321} & \metrictablesecond{0.271} & \metrictablesecond{0.289} & \metrictablesecond{0.304} & \metrictablesecond{0.274} & \metrictablesecond{0.294} & \metrictablesecond{0.221} & \metrictablesecond{0.315} \\
Ours + StableSR~\cite{wang2024exploiting} & \metrictablethird{0.248} & \metrictablebest{0.310} & \metrictablebest{0.355} & \metrictablebest{0.246} & \metrictablebest{0.228} & \metrictablebest{0.262} & \metrictablebest{0.259} & \metrictablebest{0.284} & \metrictablebest{0.285} & \metrictablebest{0.225} & \metrictablebest{0.290} & \metrictablebest{0.313} & \metrictablebest{0.257} & \metrictablebest{0.281} & \metrictablebest{0.292} & \metrictablebest{0.253} & \metrictablebest{0.272} & \metrictablebest{0.211} & \metrictablebest{0.300} \\
\bottomrule
\end{tabular}
}
\resizebox{0.6\textwidth}{!}{
\begin{tabular}{l|cc|cccccccccc}
\toprule
\multirow{2}{*}{\textbf{Method}} & \multicolumn{2}{c|}{\textbf{Deep Blending~\cite{DeepBlending2018}}} & \multicolumn{9}{c}{\textbf{Mip-NeRF 360~\cite{barron2022mipnerf360}}} \\
& \textit{drjohnson} & \textit{playroom} & \textit{bicycle} & \textit{bonsai} & \textit{counter} & \textit{flowers} & \textit{garden} & \textit{kitchen} & \textit{room} & \textit{stump} & \textit{treehill} \\
\midrule
3DGS (LR)~\cite{kerbl3Dgaussians} & 0.332 & 0.339 & 0.424 & 0.328 & 0.306 & 0.477 & 0.400 & 0.312 & 0.293 & 0.436 & 0.482 \\
3DGS~\cite{kerbl3Dgaussians} + StableSR~\cite{wang2024exploiting} & \metrictablethird{0.315} & 0.335 & 0.407 & 0.256 & 0.258 & 0.461 & 0.356 & 0.301 & 0.267 & 0.394 & 0.460 \\
Mip-Splatting~\cite{Yu2024MipSplatting} & 0.334 & \metrictablesecond{0.321} & \metrictablebest{0.318} & \metrictablebest{0.212} & \metrictablethird{0.242} & \metrictablebest{0.397} & \metrictablebest{0.292} & \metrictablesecond{0.205} & \metrictablebest{0.245} & \metrictablebest{0.314} & \metrictablebest{0.403} \\
SRGS~\cite{feng2024srgssuperresolution3dgaussian} + StableSR~\cite{wang2024exploiting} & \metrictablesecond{0.311} & \metrictablethird{0.323} & \metrictablesecond{0.349} & \metrictablesecond{0.232} & \metrictablebest{0.239} & \metrictablesecond{0.420} & \metrictablesecond{0.315} & \metrictablethird{0.249} & \metrictablesecond{0.250} & \metrictablesecond{0.368} & \metrictablesecond{0.428} \\
Ours + StableSR~\cite{wang2024exploiting} & \metrictablebest{0.300} & \metrictablebest{0.312} & \metrictablethird{0.368} & \metrictablethird{0.238} & \metrictablesecond{0.241} & \metrictablethird{0.454} & \metrictablethird{0.328} & \metrictablebest{0.202} & \metrictablethird{0.253} & \metrictablethird{0.371} & \metrictablethird{0.452} \\
\bottomrule
\end{tabular}
}
\label{tab:lpips}
\end{table*}

\begin{table*}[t]
\centering
\caption{\textbf{FID~$\downarrow$ on each scene in Tanks \& Temples~\cite{Knapitsch2017tandt}, Deep Blending~\cite{DeepBlending2018}, and Mip-NeRF 360~\cite{barron2022mipnerf360}.}}
\resizebox{\textwidth}{!}{
\begin{tabular}{l|cccccccccccccccccccc}
\toprule
\multirow{2}{*}{\textbf{Method}} & \multicolumn{19}{c}{\textbf{Tanks \& Temples~\cite{Knapitsch2017tandt}}} \\
& \textit{auditorium} & \textit{ignatius} & \textit{palace} & \textit{ballroom} & \textit{panther} & \textit{barn} & \textit{lighthouse} & \textit{playground} & \textit{courtroom} & \textit{m60} & \textit{temple} & \textit{caterpillar} & \textit{family} & \textit{train} & \textit{francis} & \textit{truck} & \textit{church} & \textit{horse} & \textit{museum} \\
\midrule
3DGS (LR)~\cite{kerbl3Dgaussians} & 86.75 & 85.58 & \metrictablesecond{83.25} & 73.89 & 40.93 & 40.48 & 84.46 & 82.51 & 99.19 & 53.16 & 81.79 & 29.93 & 113.45 & 43.80 & 29.64 & 35.69 & 58.24 & 148.48 & 88.82 \\
3DGS~\cite{kerbl3Dgaussians} + StableSR~\cite{wang2024exploiting} & \metrictablethird{68.24} & 114.03 & \metrictablethird{83.65} & 49.37 & 38.50 & 48.11 & 72.12 & 66.81 & \metrictablethird{64.95} & 40.68 & \metrictablethird{78.22} & 25.98 & 68.88 & 35.61 & 31.14 & 32.99 & 54.69 & 97.74 & \metrictablethird{54.73} \\
Mip-Splatting~\cite{Yu2024MipSplatting} & 78.68 & \metrictablesecond{52.74} & 138.29 & \metrictablethird{40.43} & \metrictablesecond{24.78} & \metrictablesecond{36.49} & \metrictablethird{67.71} & \metrictablethird{59.20} & 69.94 & \metrictablesecond{25.38} & 82.25 & \metrictablethird{20.50} & \metrictablesecond{38.03} & \metrictablethird{30.99} & \metrictablesecond{20.36} & \metrictablesecond{20.20} & \metrictablethird{48.50} & \metrictablethird{81.54} & 60.69 \\
SRGS~\cite{feng2024srgssuperresolution3dgaussian} + StableSR~\cite{wang2024exploiting} & \metrictablebest{58.94} & \metrictablethird{79.04} & 97.00 & \metrictablesecond{37.35} & \metrictablethird{26.97} & \metrictablethird{36.74} & \metrictablesecond{66.75} & \metrictablesecond{54.86} & \metrictablesecond{59.58} & \metrictablethird{27.03} & \metrictablesecond{75.46} & \metrictablesecond{20.10} & \metrictablethird{43.93} & \metrictablesecond{29.88} & \metrictablethird{23.64} & \metrictablethird{22.37} & \metrictablesecond{45.17} & \metrictablesecond{77.02} & \metrictablesecond{51.28} \\
Ours + StableSR~\cite{wang2024exploiting} & \metrictablesecond{60.48} & \metrictablebest{46.33} & \metrictablebest{70.53} & \metrictablebest{27.26} & \metrictablebest{20.96} & \metrictablebest{27.20} & \metrictablebest{57.93} & \metrictablebest{35.69} & \metrictablebest{54.34} & \metrictablebest{19.79} & \metrictablebest{63.87} & \metrictablebest{16.42} & \metrictablebest{28.28} & \metrictablebest{22.83} & \metrictablebest{16.72} & \metrictablebest{14.82} & \metrictablebest{32.61} & \metrictablebest{58.61} & \metrictablebest{41.93} \\
\bottomrule
\end{tabular}
}
\resizebox{0.6\textwidth}{!}{
\begin{tabular}{l|cc|cccccccccc}
\toprule
\multirow{2}{*}{\textbf{Method}} & \multicolumn{2}{c|}{\textbf{Deep Blending~\cite{DeepBlending2018}}} & \multicolumn{9}{c}{\textbf{Mip-NeRF 360~\cite{barron2022mipnerf360}}} \\
& \textit{drjohnson} & \textit{playroom} & \textit{bicycle} & \textit{bonsai} & \textit{counter} & \textit{flowers} & \textit{garden} & \textit{kitchen} & \textit{room} & \textit{stump} & \textit{treehill} \\
\midrule
3DGS (LR)~\cite{kerbl3Dgaussians} & 60.62 & 72.04 & 30.50 & 72.93 & 79.04 & 77.01 & 47.98 & 59.89 & 41.92 & 89.83 & 42.08 \\
3DGS~\cite{kerbl3Dgaussians} + StableSR~\cite{wang2024exploiting} & \metrictablethird{51.64} & 67.41 & 23.55 & 54.42 & 64.89 & 73.04 & 19.38 & 49.64 & 37.09 & 105.27 & 40.16 \\
Mip-Splatting~\cite{Yu2024MipSplatting} & 53.90 & \metrictablesecond{50.26} & \metrictablebest{10.21} & \metrictablebest{20.19} & \metrictablebest{36.29} & \metrictablebest{22.76} & \metrictablebest{9.05} & \metrictablesecond{12.04} & \metrictablebest{17.57} & \metrictablebest{37.03} & \metrictablebest{20.21} \\
SRGS~\cite{feng2024srgssuperresolution3dgaussian} + StableSR~\cite{wang2024exploiting} & \metrictablesecond{47.05} & \metrictablethird{51.10} & \metrictablesecond{11.37} & \metrictablethird{26.41} & \metrictablethird{41.87} & \metrictablesecond{46.01} & \metrictablesecond{10.46} & \metrictablethird{15.43} & \metrictablethird{21.62} & \metrictablesecond{42.73} & \metrictablesecond{26.52} \\
Ours + StableSR~\cite{wang2024exploiting} & \metrictablebest{43.19} & \metrictablebest{45.09} & \metrictablethird{11.79} & \metrictablesecond{22.40} & \metrictablesecond{38.03} & \metrictablethird{58.56} & \metrictablethird{10.63} & \metrictablebest{8.47} & \metrictablesecond{17.69} & \metrictablethird{50.63} & \metrictablethird{29.83} \\
\bottomrule
\end{tabular}
}
\label{tab:fid}
\end{table*}

\begin{table*}[t]
\centering
\caption{\textbf{CMMD~$\downarrow$ on each scene in Tanks \& Temples~\cite{Knapitsch2017tandt}, Deep Blending~\cite{DeepBlending2018}, and Mip-NeRF 360~\cite{barron2022mipnerf360}.}}
\resizebox{\textwidth}{!}{
\begin{tabular}{l|cccccccccccccccccccc}
\toprule
\multirow{2}{*}{\textbf{Method}} & \multicolumn{19}{c}{\textbf{Tanks \& Temples~\cite{Knapitsch2017tandt}}} \\
& \textit{auditorium} & \textit{ignatius} & \textit{palace} & \textit{ballroom} & \textit{panther} & \textit{barn} & \textit{lighthouse} & \textit{playground} & \textit{courtroom} & \textit{m60} & \textit{temple} & \textit{caterpillar} & \textit{family} & \textit{train} & \textit{francis} & \textit{truck} & \textit{church} & \textit{horse} & \textit{museum} \\
\midrule
3DGS (LR)~\cite{kerbl3Dgaussians} & 1.770 & 2.234 & 2.361 & 1.602 & 0.718 & 1.699 & 1.540 & 1.925 & 1.616 & 1.202 & 2.522 & 1.897 & 3.038 & 2.153 & 2.758 & 2.140 & 1.705 & 3.211 & 2.165 \\
3DGS~\cite{kerbl3Dgaussians} + StableSR~\cite{wang2024exploiting} & \metrictablethird{1.093} & \metrictablesecond{1.032} & \metrictablebest{1.491} & \metrictablethird{1.073} & 0.791 & \metrictablesecond{1.068} & 1.092 & \metrictablethird{1.139} & \metrictablethird{0.987} & 0.907 & \metrictablebest{1.597} & 1.464 & \metrictablethird{0.647} & \metrictablebest{1.330} & \metrictablethird{1.401} & 1.143 & 1.163 & \metrictablesecond{0.779} & \metrictablebest{1.142} \\
Mip-Splatting~\cite{Yu2024MipSplatting} & 1.246 & \metrictablethird{1.046} & 1.624 & \metrictablesecond{0.787} & \metrictablethird{0.633} & 1.140 & \metrictablethird{0.994} & 1.151 & 1.024 & \metrictablethird{0.579} & 1.853 & \metrictablebest{1.243} & 0.863 & \metrictablethird{1.384} & \metrictablebest{1.286} & \metrictablesecond{1.080} & \metrictablesecond{1.029} & 1.105 & 1.528 \\
SRGS~\cite{feng2024srgssuperresolution3dgaussian} + StableSR~\cite{wang2024exploiting} & \metrictablesecond{1.066} & \metrictablebest{0.931} & \metrictablethird{1.582} & \metrictablesecond{0.787} & \metrictablesecond{0.516} & \metrictablebest{0.942} & \metrictablesecond{0.978} & \metrictablebest{0.998} & \metrictablebest{0.943} & \metrictablesecond{0.523} & \metrictablethird{1.795} & \metrictablesecond{1.261} & \metrictablesecond{0.557} & \metrictablesecond{1.383} & 1.408 & \metrictablethird{1.140} & \metrictablethird{1.049} & \metrictablebest{0.658} & \metrictablethird{1.400} \\
Ours + StableSR~\cite{wang2024exploiting} & \metrictablebest{0.993} & 1.121 & \metrictablesecond{1.568} & \metrictablebest{0.643} & \metrictablebest{0.351} & \metrictablethird{1.134} & \metrictablebest{0.901} & \metrictablesecond{1.025} & \metrictablesecond{0.951} & \metrictablebest{0.443} & \metrictablesecond{1.699} & \metrictablethird{1.330} & \metrictablebest{0.521} & 1.529 & \metrictablesecond{1.355} & \metrictablebest{1.061} & \metrictablebest{0.944} & \metrictablethird{0.897} & \metrictablesecond{1.286} \\
\bottomrule
\end{tabular}
}
\resizebox{0.6\textwidth}{!}{
\begin{tabular}{l|cc|cccccccccc}
\toprule
\multirow{2}{*}{\textbf{Method}} & \multicolumn{2}{c|}{\textbf{Deep Blending~\cite{DeepBlending2018}}} & \multicolumn{9}{c}{\textbf{Mip-NeRF 360~\cite{barron2022mipnerf360}}} \\
& \textit{drjohnson} & \textit{playroom} & \textit{bicycle} & \textit{bonsai} & \textit{counter} & \textit{flowers} & \textit{garden} & \textit{kitchen} & \textit{room} & \textit{stump} & \textit{treehill} \\
\midrule
3DGS (LR)~\cite{kerbl3Dgaussians} & 1.003 & 0.733 & 0.392 & 1.011 & 0.319 & 1.354 & 0.242 & 0.346 & \metrictablethird{0.261} & 0.926 & 0.956 \\
3DGS~\cite{kerbl3Dgaussians} + StableSR~\cite{wang2024exploiting} & \metrictablethird{0.807} & 0.714 & 0.621 & 0.934 & 0.440 & 1.409 & 0.310 & 0.441 & 0.578 & 0.767 & 1.023 \\
Mip-Splatting~\cite{Yu2024MipSplatting} & 0.828 & \metrictablethird{0.552} & \metrictablebest{0.099} & \metrictablesecond{0.346} & \metrictablesecond{0.195} & \metrictablebest{0.215} & \metrictablebest{0.036} & \metrictablesecond{0.191} & \metrictablesecond{0.187} & \metrictablebest{0.125} & \metrictablebest{0.257} \\
SRGS~\cite{feng2024srgssuperresolution3dgaussian} + StableSR~\cite{wang2024exploiting} & \metrictablesecond{0.715} & \metrictablesecond{0.546} & \metrictablesecond{0.138} & \metrictablethird{0.417} & \metrictablethird{0.279} & \metrictablesecond{0.782} & \metrictablethird{0.081} & \metrictablethird{0.228} & 0.378 & \metrictablesecond{0.261} & \metrictablesecond{0.486} \\
Ours + StableSR~\cite{wang2024exploiting} & \metrictablebest{0.558} & \metrictablebest{0.434} & \metrictablethird{0.159} & \metrictablebest{0.266} & \metrictablebest{0.139} & \metrictablethird{1.108} & \metrictablesecond{0.065} & \metrictablebest{0.119} & \metrictablebest{0.184} & \metrictablethird{0.406} & \metrictablethird{0.610} \\
\bottomrule
\end{tabular}
}
\label{tab:cmmd}
\end{table*}

\begin{table*}[t]
\centering
\caption{\textbf{DreamSim~$\downarrow$ on each scene in Tanks \& Temples~\cite{Knapitsch2017tandt}, Deep Blending~\cite{DeepBlending2018}, and Mip-NeRF 360~\cite{barron2022mipnerf360}.}}
\resizebox{\textwidth}{!}{
\begin{tabular}{l|cccccccccccccccccccc}
\toprule
\multirow{2}{*}{\textbf{Method}} & \multicolumn{19}{c}{\textbf{Tanks \& Temples~\cite{Knapitsch2017tandt}}} \\
& \textit{auditorium} & \textit{ignatius} & \textit{palace} & \textit{ballroom} & \textit{panther} & \textit{barn} & \textit{lighthouse} & \textit{playground} & \textit{courtroom} & \textit{m60} & \textit{temple} & \textit{caterpillar} & \textit{family} & \textit{train} & \textit{francis} & \textit{truck} & \textit{church} & \textit{horse} & \textit{museum} \\
\midrule
3DGS (LR)~\cite{kerbl3Dgaussians} & 0.0969 & 0.0884 & \metrictablethird{0.1548} & 0.0881 & 0.0399 & 0.0549 & 0.1154 & 0.0806 & 0.1074 & 0.0552 & 0.1156 & 0.0595 & 0.1215 & 0.0747 & 0.0710 & 0.0620 & 0.0795 & 0.1142 & 0.1207 \\
3DGS~\cite{kerbl3Dgaussians} + StableSR~\cite{wang2024exploiting} & \metrictablethird{0.0634} & 0.0703 & \metrictablesecond{0.1497} & 0.0438 & 0.0443 & 0.0613 & 0.0909 & 0.0759 & \metrictablethird{0.0566} & 0.0591 & \metrictablethird{0.0955} & 0.0516 & 0.0565 & 0.0652 & 0.0555 & 0.0590 & 0.0558 & 0.0505 & \metrictablethird{0.0627} \\
Mip-Splatting~\cite{Yu2024MipSplatting} & 0.0739 & \metrictablesecond{0.0465} & 0.2459 & \metrictablethird{0.0312} & \metrictablesecond{0.0216} & \metrictablesecond{0.0392} & \metrictablethird{0.0887} & \metrictablesecond{0.0541} & 0.0592 & \metrictablesecond{0.0266} & 0.1001 & \metrictablesecond{0.0392} & \metrictablesecond{0.0278} & \metrictablesecond{0.0479} & \metrictablesecond{0.0341} & \metrictablesecond{0.0293} & \metrictablethird{0.0541} & \metrictablethird{0.0438} & 0.0703 \\
SRGS~\cite{feng2024srgssuperresolution3dgaussian} + StableSR~\cite{wang2024exploiting} & \metrictablesecond{0.0500} & \metrictablethird{0.0511} & 0.1603 & \metrictablesecond{0.0271} & \metrictablethird{0.0264} & \metrictablethird{0.0399} & \metrictablesecond{0.0792} & \metrictablethird{0.0564} & \metrictablesecond{0.0466} & \metrictablethird{0.0369} & \metrictablesecond{0.0936} & \metrictablethird{0.0416} & \metrictablethird{0.0359} & \metrictablethird{0.0504} & \metrictablethird{0.0402} & \metrictablethird{0.0402} & \metrictablesecond{0.0459} & \metrictablesecond{0.0390} & \metrictablesecond{0.0567} \\
Ours + StableSR~\cite{wang2024exploiting} & \metrictablebest{0.0490} & \metrictablebest{0.0381} & \metrictablebest{0.1272} & \metrictablebest{0.0202} & \metrictablebest{0.0187} & \metrictablebest{0.0314} & \metrictablebest{0.0625} & \metrictablebest{0.0369} & \metrictablebest{0.0422} & \metrictablebest{0.0239} & \metrictablebest{0.0720} & \metrictablebest{0.0315} & \metrictablebest{0.0229} & \metrictablebest{0.0410} & \metrictablebest{0.0309} & \metrictablebest{0.0268} & \metrictablebest{0.0331} & \metrictablebest{0.0279} & \metrictablebest{0.0486} \\
\bottomrule
\end{tabular}
}
\resizebox{0.6\textwidth}{!}{
\begin{tabular}{l|cc|cccccccccc}
\toprule
\multirow{2}{*}{\textbf{Method}} & \multicolumn{2}{c|}{\textbf{Deep Blending~\cite{DeepBlending2018}}} & \multicolumn{9}{c}{\textbf{Mip-NeRF 360~\cite{barron2022mipnerf360}}} \\
& \textit{drjohnson} & \textit{playroom} & \textit{bicycle} & \textit{bonsai} & \textit{counter} & \textit{flowers} & \textit{garden} & \textit{kitchen} & \textit{room} & \textit{stump} & \textit{treehill} \\
\midrule
3DGS (LR)~\cite{kerbl3Dgaussians} & 0.0613 & 0.0534 & 0.0624 & 0.0659 & 0.0453 & 0.0562 & 0.0452 & 0.0454 & 0.0322 & 0.0586 & 0.0558 \\
3DGS~\cite{kerbl3Dgaussians} + StableSR~\cite{wang2024exploiting} & 0.0499 & 0.0510 & 0.0348 & 0.0320 & 0.0377 & 0.0598 & 0.0170 & 0.0240 & 0.0345 & 0.0536 & 0.0553 \\
Mip-Splatting~\cite{Yu2024MipSplatting} & \metrictablethird{0.0451} & \metrictablesecond{0.0345} & \metrictablebest{0.0126} & \metrictablebest{0.0122} & \metrictablebest{0.0166} & \metrictablebest{0.0124} & \metrictablebest{0.0070} & \metrictablesecond{0.0046} & \metrictablesecond{0.0163} & \metrictablebest{0.0117} & \metrictablebest{0.0252} \\
SRGS~\cite{feng2024srgssuperresolution3dgaussian} + StableSR~\cite{wang2024exploiting} & \metrictablesecond{0.0439} & \metrictablethird{0.0378} & \metrictablesecond{0.0150} & \metrictablethird{0.0160} & \metrictablethird{0.0226} & \metrictablesecond{0.0331} & \metrictablethird{0.0079} & \metrictablethird{0.0066} & \metrictablethird{0.0212} & \metrictablethird{0.0178} & \metrictablesecond{0.0345} \\
Ours + StableSR~\cite{wang2024exploiting} & \metrictablebest{0.0332} & \metrictablebest{0.0327} & \metrictablethird{0.0151} & \metrictablesecond{0.0123} & \metrictablesecond{0.0177} & \metrictablethird{0.0380} & \metrictablesecond{0.0071} & \metrictablebest{0.0037} & \metrictablebest{0.0144} & \metrictablesecond{0.0168} & \metrictablethird{0.0364} \\
\bottomrule
\end{tabular}
}
\label{tab:dreamsim}
\end{table*}

\begin{table*}[t]
\centering
\caption{\textbf{MUSIQ~$\uparrow$ on each scene in Tanks \& Temples~\cite{Knapitsch2017tandt}, Deep Blending~\cite{DeepBlending2018}, and Mip-NeRF 360~\cite{barron2022mipnerf360}.}}
\resizebox{\textwidth}{!}{
\begin{tabular}{l|cccccccccccccccccccc}
\toprule
\multirow{2}{*}{\textbf{Method}} & \multicolumn{19}{c}{\textbf{Tanks \& Temples~\cite{Knapitsch2017tandt}}} \\
& \textit{auditorium} & \textit{ignatius} & \textit{palace} & \textit{ballroom} & \textit{panther} & \textit{barn} & \textit{lighthouse} & \textit{playground} & \textit{courtroom} & \textit{m60} & \textit{temple} & \textit{caterpillar} & \textit{family} & \textit{train} & \textit{francis} & \textit{truck} & \textit{church} & \textit{horse} & \textit{museum} \\
\midrule
3DGS (LR)~\cite{kerbl3Dgaussians} & 55.443 & \metrictablebest{64.965} & \metrictablebest{51.161} & \metrictablethird{56.309} & 54.610 & \metrictablethird{60.048} & \metrictablebest{48.785} & \metrictablesecond{54.510} & \metrictablesecond{55.349} & 53.631 & \metrictablebest{57.565} & \metrictablethird{59.116} & \metrictablebest{65.987} & \metrictablesecond{60.573} & \metrictablebest{59.605} & \metrictablesecond{61.675} & \metrictablebest{57.991} & \metrictablesecond{57.572} & \metrictablesecond{62.845} \\
3DGS~\cite{kerbl3Dgaussians} + StableSR~\cite{wang2024exploiting} & \metrictablebest{58.573} & \metrictablethird{58.466} & \metrictablethird{49.326} & \metrictablesecond{56.501} & \metrictablebest{57.051} & \metrictablesecond{60.822} & \metrictablethird{46.028} & \metrictablethird{53.817} & \metrictablethird{55.209} & \metrictablesecond{54.985} & \metrictablethird{54.748} & \metrictablesecond{60.520} & \metrictablethird{62.883} & \metrictablebest{60.841} & \metrictablethird{55.264} & \metrictablethird{60.637} & \metrictablethird{55.297} & \metrictablethird{54.926} & \metrictablethird{62.311} \\
Mip-Splatting~\cite{Yu2024MipSplatting} & 51.773 & 48.745 & 42.653 & 50.651 & 46.800 & 49.040 & 33.568 & 41.815 & 45.084 & 47.621 & 46.236 & 51.797 & 49.461 & 51.300 & 40.700 & 47.208 & 47.433 & 39.068 & 53.893 \\
SRGS~\cite{feng2024srgssuperresolution3dgaussian} + StableSR~\cite{wang2024exploiting} & \metrictablethird{57.276} & 57.690 & 48.339 & 54.321 & \metrictablethird{54.969} & 58.305 & 44.228 & 52.796 & 54.435 & \metrictablethird{53.662} & 52.762 & 58.732 & 61.687 & 58.721 & 54.220 & 57.781 & 54.378 & 53.369 & 61.291 \\
Ours + StableSR~\cite{wang2024exploiting} & \metrictablesecond{58.504} & \metrictablesecond{60.078} & \metrictablesecond{50.762} & \metrictablebest{59.501} & \metrictablesecond{56.915} & \metrictablebest{61.660} & \metrictablesecond{47.882} & \metrictablebest{58.036} & \metrictablebest{57.351} & \metrictablebest{55.500} & \metrictablesecond{55.978} & \metrictablebest{60.757} & \metrictablesecond{65.967} & \metrictablethird{60.214} & \metrictablesecond{57.281} & \metrictablebest{62.616} & \metrictablesecond{57.574} & \metrictablebest{58.244} & \metrictablebest{63.491} \\
\bottomrule
\end{tabular}
}
\resizebox{0.6\textwidth}{!}{
\begin{tabular}{l|cc|cccccccccc}
\toprule
\multirow{2}{*}{\textbf{Method}} & \multicolumn{2}{c|}{\textbf{Deep Blending~\cite{DeepBlending2018}}} & \multicolumn{9}{c}{\textbf{Mip-NeRF 360~\cite{barron2022mipnerf360}}} \\
& \textit{drjohnson} & \textit{playroom} & \textit{bicycle} & \textit{bonsai} & \textit{counter} & \textit{flowers} & \textit{garden} & \textit{kitchen} & \textit{room} & \textit{stump} & \textit{treehill} \\
\midrule
3DGS (LR)~\cite{kerbl3Dgaussians} & 47.916 & 49.425 & 54.239 & 56.108 & \metrictablethird{55.829} & \metrictablebest{58.968} & \metrictablebest{58.961} & \metrictablesecond{63.552} & 50.931 & \metrictablebest{52.085} & \metrictablebest{41.783} \\
3DGS~\cite{kerbl3Dgaussians} + StableSR~\cite{wang2024exploiting} & \metrictablebest{52.919} & \metrictablesecond{53.324} & \metrictablesecond{55.767} & \metrictablebest{59.725} & \metrictablebest{57.965} & \metrictablethird{52.194} & \metrictablethird{54.576} & \metrictablethird{59.952} & \metrictablebest{58.135} & \metrictablethird{44.337} & \metrictablesecond{38.276} \\
Mip-Splatting~\cite{Yu2024MipSplatting} & 38.800 & 44.105 & 38.243 & 44.656 & 42.445 & 35.728 & 44.342 & 50.600 & 41.878 & 35.866 & 35.112 \\
SRGS~\cite{feng2024srgssuperresolution3dgaussian} + StableSR~\cite{wang2024exploiting} & \metrictablesecond{50.849} & \metrictablethird{52.035} & \metrictablethird{54.818} & \metrictablethird{56.237} & 55.391 & \metrictablesecond{53.542} & 54.318 & 58.321 & \metrictablesecond{55.637} & 44.039 & 37.260 \\
Ours + StableSR~\cite{wang2024exploiting} & \metrictablethird{50.594} & \metrictablebest{55.444} & \metrictablebest{57.923} & \metrictablesecond{58.501} & \metrictablesecond{57.752} & 51.206 & \metrictablesecond{56.236} & \metrictablebest{64.952} & \metrictablethird{55.154} & \metrictablesecond{49.551} & \metrictablethird{38.014} \\
\bottomrule
\end{tabular}
}
\label{tab:musiq}
\end{table*}

\begin{table*}[t]
\centering
\caption{\textbf{NIQE~$\downarrow$ on each scene in Tanks \& Temples~\cite{Knapitsch2017tandt}, Deep Blending~\cite{DeepBlending2018}, and Mip-NeRF 360~\cite{barron2022mipnerf360}.}}
\resizebox{\textwidth}{!}{
\begin{tabular}{l|cccccccccccccccccccc}
\toprule
\multirow{2}{*}{\textbf{Method}} & \multicolumn{19}{c}{\textbf{Tanks \& Temples~\cite{Knapitsch2017tandt}}} \\
& \textit{auditorium} & \textit{ignatius} & \textit{palace} & \textit{ballroom} & \textit{panther} & \textit{barn} & \textit{lighthouse} & \textit{playground} & \textit{courtroom} & \textit{m60} & \textit{temple} & \textit{caterpillar} & \textit{family} & \textit{train} & \textit{francis} & \textit{truck} & \textit{church} & \textit{horse} & \textit{museum} \\
\midrule
3DGS (LR)~\cite{kerbl3Dgaussians} & \metrictablebest{4.718} & \metrictablebest{2.628} & \metrictablebest{4.293} & \metrictablebest{2.791} & \metrictablebest{3.884} & \metrictablebest{3.944} & \metrictablebest{4.097} & \metrictablebest{2.744} & \metrictablebest{3.385} & \metrictablebest{3.625} & \metrictablebest{4.081} & \metrictablebest{2.873} & \metrictablebest{2.605} & \metrictablebest{2.994} & \metrictablebest{4.269} & \metrictablebest{2.784} & \metrictablebest{3.169} & \metrictablebest{3.011} & \metrictablebest{2.926} \\
3DGS~\cite{kerbl3Dgaussians} + StableSR~\cite{wang2024exploiting} & 6.531 & 4.545 & 5.820 & 4.335 & 5.239 & 5.229 & 5.415 & 4.991 & 4.830 & 5.022 & 4.844 & 4.690 & 4.214 & 4.363 & 5.611 & 4.361 & 5.005 & 4.564 & 4.356 \\
Mip-Splatting~\cite{Yu2024MipSplatting} & \metrictablethird{5.482} & 4.693 & 6.567 & 4.013 & 5.614 & 5.302 & 5.482 & 5.086 & 4.569 & 4.955 & 5.039 & 4.632 & 5.077 & 4.449 & 5.982 & 5.022 & 4.653 & 5.027 & 4.179 \\
SRGS~\cite{feng2024srgssuperresolution3dgaussian} + StableSR~\cite{wang2024exploiting} & 6.128 & \metrictablethird{4.187} & \metrictablethird{5.629} & \metrictablethird{3.978} & \metrictablethird{4.904} & \metrictablethird{4.988} & \metrictablethird{5.025} & \metrictablethird{4.537} & \metrictablethird{4.562} & \metrictablethird{4.716} & \metrictablethird{4.573} & \metrictablethird{4.323} & \metrictablethird{4.026} & \metrictablethird{4.137} & \metrictablethird{5.234} & \metrictablethird{4.096} & \metrictablethird{4.644} & \metrictablethird{4.280} & \metrictablethird{4.050} \\
Ours + StableSR~\cite{wang2024exploiting} & \metrictablesecond{5.330} & \metrictablesecond{3.440} & \metrictablesecond{4.720} & \metrictablesecond{3.318} & \metrictablesecond{3.997} & \metrictablesecond{4.657} & \metrictablesecond{4.309} & \metrictablesecond{3.644} & \metrictablesecond{4.029} & \metrictablesecond{3.765} & \metrictablesecond{4.167} & \metrictablesecond{3.340} & \metrictablesecond{3.435} & \metrictablesecond{3.498} & \metrictablesecond{4.559} & \metrictablesecond{3.416} & \metrictablesecond{3.785} & \metrictablesecond{3.632} & \metrictablesecond{3.582} \\
\bottomrule
\end{tabular}
}
\resizebox{0.6\textwidth}{!}{
\begin{tabular}{l|cc|cccccccccc}
\toprule
\multirow{2}{*}{\textbf{Method}} & \multicolumn{2}{c|}{\textbf{Deep Blending~\cite{DeepBlending2018}}} & \multicolumn{9}{c}{\textbf{Mip-NeRF 360~\cite{barron2022mipnerf360}}} \\
& \textit{drjohnson} & \textit{playroom} & \textit{bicycle} & \textit{bonsai} & \textit{counter} & \textit{flowers} & \textit{garden} & \textit{kitchen} & \textit{room} & \textit{stump} & \textit{treehill} \\
\midrule
3DGS (LR)~\cite{kerbl3Dgaussians} & \metrictablebest{4.760} & \metrictablebest{5.329} & \metrictablebest{2.808} & \metrictablebest{3.964} & \metrictablebest{3.535} & \metrictablebest{2.915} & \metrictablebest{2.741} & \metrictablebest{2.986} & \metrictablebest{4.210} & \metrictablebest{3.137} & \metrictablebest{2.744} \\
3DGS~\cite{kerbl3Dgaussians} + StableSR~\cite{wang2024exploiting} & 5.231 & 6.225 & 4.333 & 5.452 & 5.533 & 4.938 & 5.516 & 5.633 & 6.013 & 4.919 & 5.105 \\
Mip-Splatting~\cite{Yu2024MipSplatting} & 6.097 & 6.426 & 5.665 & 6.458 & 5.730 & 5.965 & 6.155 & 5.600 & 5.885 & 5.623 & 5.462 \\
SRGS~\cite{feng2024srgssuperresolution3dgaussian} + StableSR~\cite{wang2024exploiting} & \metrictablethird{5.208} & \metrictablethird{6.188} & \metrictablethird{4.131} & \metrictablethird{5.228} & \metrictablethird{5.214} & \metrictablethird{4.521} & \metrictablethird{5.074} & \metrictablethird{5.546} & \metrictablethird{5.840} & \metrictablethird{4.846} & \metrictablethird{4.381} \\
Ours + StableSR~\cite{wang2024exploiting} & \metrictablesecond{5.054} & \metrictablesecond{5.768} & \metrictablesecond{3.378} & \metrictablesecond{4.309} & \metrictablesecond{3.828} & \metrictablesecond{3.825} & \metrictablesecond{3.949} & \metrictablesecond{3.828} & \metrictablesecond{4.531} & \metrictablesecond{4.052} & \metrictablesecond{3.702} \\
\bottomrule
\end{tabular}
}
\label{tab:niqe}
\end{table*}

\end{document}